\documentclass[lettersize,journal]{IEEEtran}

\usepackage[OT1]{fontenc} 
\usepackage{adjustbox}
\usepackage[ruled, linesnumbered]{algorithm2e}
\usepackage{amsfonts}       
\usepackage{amsmath}
\usepackage{amssymb}
\usepackage{amsthm}
\usepackage{arydshln}
\usepackage{balance}
\usepackage{bbm}
\usepackage{bm}
\usepackage{booktabs}       
\usepackage{array}
\usepackage[font=small]{caption}
\usepackage{color}
\usepackage[noadjust]{cite}

\usepackage{enumitem} 
\usepackage{float}
\usepackage{grffile} 
\usepackage{graphicx}
\usepackage{graphics}
\usepackage{microtype}      
\usepackage{multirow}
\usepackage[caption=false,font=normalsize,labelfont=sf,textfont=sf]{subfig}
\usepackage{url}            
\usepackage[table]{xcolor} 
\usepackage{xparse}
\usepackage{colortbl}
\usepackage{threeparttable}

\ExplSyntaxOn
\NewDocumentCommand{\longdash}{ O{2} }
 {
  --\prg_replicate:nn { #1 - 1 } { \negthinspace -- }
 }
\ExplSyntaxOff

\usepackage{hyperref}       

\newcommand{\revision}[1]{\textcolor{black}{#1}}

\theoremstyle{definition}

\newtheorem{remark}{Remark}

\usepackage{shortex}

\graphicspath{{images/}} 

\allowdisplaybreaks 

\begin{document}
	

 	\title{CCDM: Continuous Conditional Diffusion Models for Image Generation}

	\author{Xin~Ding,~\IEEEmembership{Member,~IEEE,}
		Yongwei Wang,~\IEEEmembership{Member,~IEEE,}
		Kao Zhang,
		and~Z.~Jane~Wang,~\IEEEmembership{Fellow,~IEEE}
		\thanks{Xin Ding and Kao Zhang are affiliated with the School of Artificial Intelligence/School of Future Technology, Nanjing University of Information Science \& Technology (NUIST), Nanjing, China (e-mail: \{dingxin, kaozhang\}@nuist.edu.cn). They are also affiliated with the Perceptual and Generative AI Lab (PGAI Lab) at NUIST.} 
		\thanks{Yongwei Wang (corresponding author) is with Zhejiang University, China (e-mail: yongwei.wang@zju.edu.cn).}
		\thanks{Z.\ Jane Wang is with the Department of Electrical and Computer Engineering, University of British Columbia, Vancouver, Canada (e-mail: zjanew@ece.ubc.ca).}
		\thanks{Manuscript received February XX, 2025; revised February XX, 2025.} 
	}
	\markboth{Journal of \LaTeX\ Class Files,~Vol.~14, No.~8, August~2021}%
	{Shell \MakeLowercase{\textit{et al.}}: A Sample Article Using IEEEtran.cls for IEEE Journals}

	\maketitle

	\begin{abstract}
		
	    \textit{Continuous Conditional Generative Modeling} (CCGM) estimates high-dimensional data distributions, such as images, conditioned on scalar continuous variables (aka regression labels). While \textit{Continuous Conditional Generative Adversarial Networks} (CcGANs) were designed for this task, their instability during adversarial learning often leads to suboptimal results. \textit{Conditional Diffusion Models} (CDMs) offer a promising alternative, generating more realistic images, but their diffusion processes, label conditioning, and model fitting procedures are either not optimized for or incompatible with CCGM, making it difficult to integrate CcGANs' vicinal approach. To address these issues, we introduce \textit{Continuous Conditional Diffusion Models} (CCDMs), the first CDM specifically tailored for CCGM. CCDMs address existing limitations with specially designed conditional diffusion processes, a novel hard vicinal image denoising loss, a customized label embedding method, and efficient conditional sampling procedures. Through comprehensive experiments on four datasets with resolutions ranging from $64\times 64$ to $192\times 192$, we demonstrate that CCDMs outperform state-of-the-art CCGM models, establishing a new benchmark. Ablation studies further validate the model design and implementation, highlighting that some widely used CDM implementations are ineffective for the CCGM task. Our code is publicly available at \url{https://github.com/UBCDingXin/CCDM}.

	\end{abstract}
	

	\section{Introduction}\label{sec:introduction}
	
\begin{figure}[!htbp] 
		\centering
		\includegraphics[width=1\linewidth]{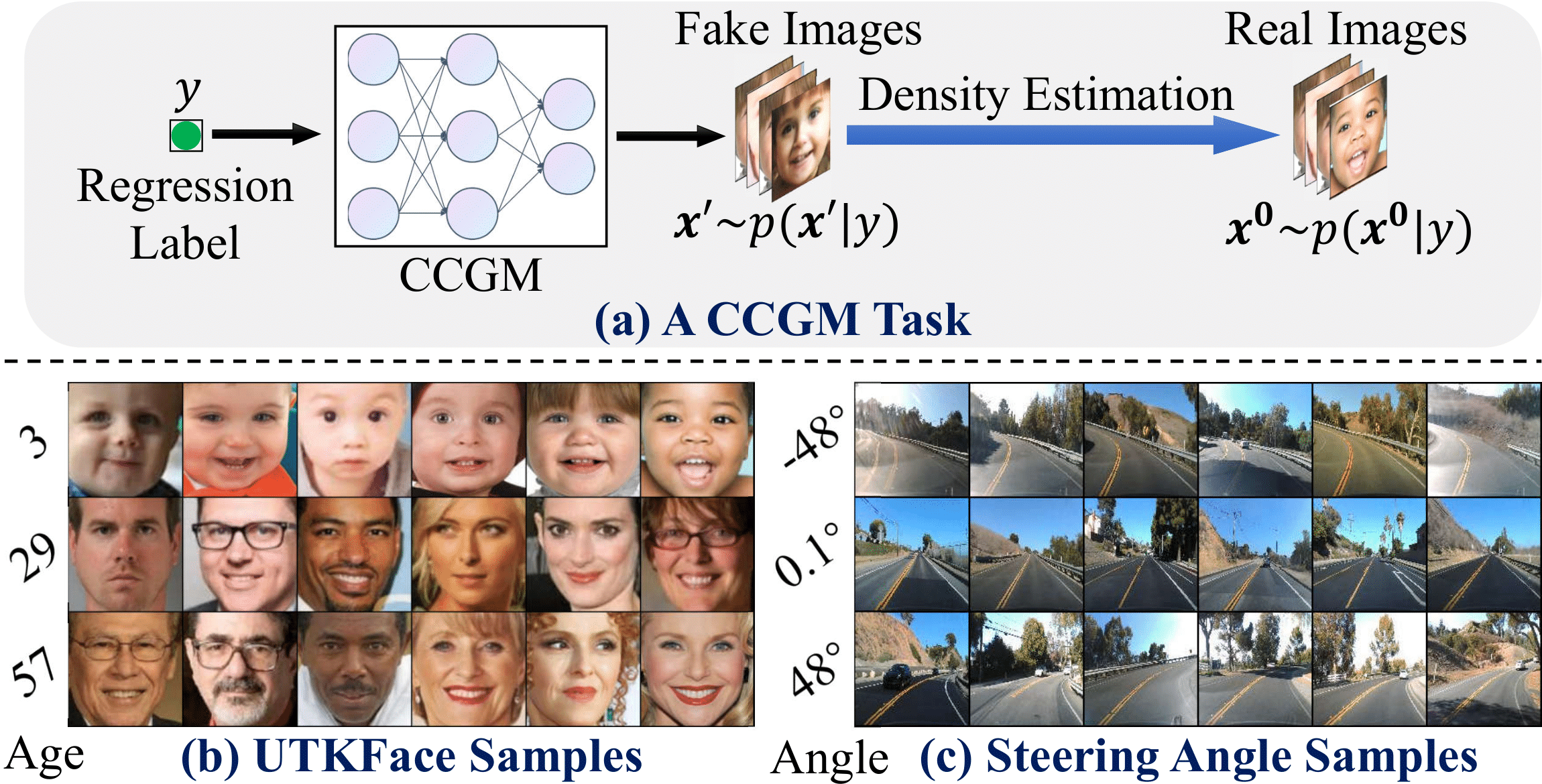}
		\caption{Illustration of the CCGM task with sample images from the UTKFace and Steering Angle datasets. }
		\label{fig:illustrative_CCGM}
	\end{figure}

	\textit{Continuous Conditional Generative Modeling} (CCGM), as depicted in Fig.~\ref{fig:illustrative_CCGM}, aims to model the probability distribution of images conditioned on scalar continuous variables. These variables, commonly referred to as regression labels, can encompass a diverse range of values, including \revision{angles, ages, counts, temperatures, coordinates, and more.} However, CCGM faces significant challenges, particularly due to the scarcity of training samples for specific regression labels and the lack of an effective label input mechanism. These limitations adversely impact the performance of conventional \textit{Conditional Generative Adversarial Networks} (cGANs) and \textit{Conditional Diffusion Models} (CDMs), as demonstrated in \cite{ding2023ccgan, ding2024turning}.
	
	As a recent advancement, Ding et al. introduced the first feasible model to the CCGM task, termed \textit{Continuous Conditional Generative Adversarial Networks} (CcGANs)~\cite{ding2023ccgan}. CcGANs address the challenges of the CCGM task by developing novel vicinal discriminator losses and label input mechanisms. Consequently, CcGANs have been extensively applied across various domains requiring precise control over generative modeling of high-dimensional data. These applications comprise engineering inverse design \cite{heyrani2021pcdgan, zhao2024ccdpm}, data augmentation for hyperspectral imaging~\cite{zhu2023data}, remote sensing image processing~\cite{giry2022sar}, model compression~\cite{ding2023distilling}, controllable point cloud generation~\cite{triess2022point}, and more. Despite their success, CcGANs still face crucial limitations~\cite{ding2024turning}, e.g., occasionally producing low-quality images marked by poor visual fidelity and label inconsistency—instances where generated images deviate from the specified conditions.
	
	Diffusion models now lead image generation, consistently outperforming GANs in quality \cite{dhariwal2021diffusion, ho2021classifier, rombach2022high, peebles2023scalable, esser2024scaling, labs2025flux1kontextflowmatching}, making their adaptation to CCGM a promising path to surpass CcGANs \cite{ding2023ccgan, ding2024turning}. However, existing \textit{Conditional Diffusion Models} (CDMs) are designed for discrete conditions like class labels \cite{dhariwal2021diffusion, ho2021classifier} or text descriptions \cite{rombach2022high, peebles2023scalable, esser2024scaling, li2024g, qu2024discriminative, labs2025flux1kontextflowmatching}, making them unsuitable for regression labels in CCGM due to numerous limitations: their diffusion processes lack an explicit condition incorporation mechanism, leading to suboptimal performance, particularly for label inconsistency and poor visual fidelity; their label input mechanisms require modification for regression labels; the commonly used noise prediction loss and hyperparameter selection schemes are largely ineffective in CCGM; and the soft vicinal loss optimal for CcGANs does not translate well to CDMs. Consequently, integrating CDMs with CcGAN techniques faces significant challenges, underscoring an urgent need for developing novel diffusion processes, loss functions, label input mechanisms, and training/sampling strategies specially tailored to CCGM.
	
	Motivated by the aforementioned challenges, we introduce \textit{Continuous Conditional Diffusion Models} (CCDMs) in this paper—a novel framework designed to overcome the crucial limitations of existing CDMs and offer a comprehensive guideline for their application in the CCGM task. Our key contributions are summarized as follows:
	
	\begin{itemize}		
		
		\item To improve control over generated images and enhance overall generation quality, we propose novel conditional diffusion processes that explicitly integrate conditional information through $y$-dependent state transitions, as detailed in Section~\ref{sec:conditional_diffusion_process}.
		
		\item To address the data sparsity issue in CCGM, we derive the \textit{Hard Vicinal Negative Log-Likelihood} (HV-NLL), inspired by the vicinal technique from CcGANs~\cite{ding2023ccgan}. Building on HV-NLL and the modified diffusion, we introduce the \textit{Hard Vicinal Image Denoising Loss} (HVIDL) in Section~\ref{sec:model_fitting}, a novel training loss that incorporates a hard vicinal weight and a $y$-dependent covariance matrix. We also argue that the widely used noise prediction loss~\cite{ho2020denoising} for diffusion models and the soft vicinal loss, while optimal for CcGANs, are unsuitable for CCGM—a claim supported by ablation studies in Section~\ref{sec:ablation}.
		
		\item \revision{In Section~\ref{sec:conditional_sampling}, we employ a multi-step sampling algorithm optimized for CCGM, with configurations distinct from class-conditional or text-to-image generation.} In Section~\ref{sec:dmd_sampling}, we introduce DMD2-M, a one-step sampler inspired by DMD2~\cite{yin2024improved}, enhanced with advanced GAN architectures and training techniques to adapt it for the CCGM task. DMD2-M significantly accelerates sampling, with a slight trade-off in image quality.
		
		\item Building on the comprehensive experiments in Sections~\ref{sec:experiment} and \ref{sec:ablation}, we demonstrate that CCDM achieves state-of-the-art performance. Additionally, we offer a thorough implementation guideline for CCDM.
		
	\end{itemize}
	
	It is worth mentioning that the proposed CCDM substantially differs from a recent work CcDPM~\cite{zhao2024ccdpm} in four aspects:
	\begin{itemize}
		\item CcDPM focuses on \textbf{non-image} data, whereas high-dimensional image generation is significantly more challenging than non-image ones. Also, CcDPM was designed for \textbf{multi-dimensional} labels, lacking an efficient mechanism for scalar regression labels, whereas our CCDMs are specifically developed for \textbf{image} data with \textbf{scalar} continuous labels, making CCDM the \textit{\textbf{first CDM tailored for the CCGM task}}.
		
		\item CcDPM employs a noise prediction loss weighted by soft vicinity, while CCDM proposes a novel image denoising loss weighted by hard vicinity, which our ablation study in Section~\ref{sec:ablation} demonstrates significantly better performance in CCGM tasks.
		
		\item CcDPM's diffusion processes are not directly tied to regression labels, whereas CCDM explicitly integrates regression labels into both forward and reverse processes, enhancing image quality.
		
		\item CCDM provides a mathematically rigorous derivation of its modified diffusion processes and training loss in Appendix~\ref{sec:supp_detailed_derivations}, which is absent in CcDPM. These distinctions underscore CCDM's unique effectiveness in image generation with scalar continuous labels.
	\end{itemize}

	\section{Related Work}\label{sec:related_work}

	\subsection{Continuous Conditional GANs} 
	
	CcGANs, introduced by Ding et al.~\cite{ding2021ccgan, ding2023ccgan}, represent the pioneering approach to the \textit{Continuous Conditional Generative Modeling} (CCGM) task. As shown in Fig.~\ref{fig:illustrative_CCGM}, mathematically, CcGANs are devised to estimate the probability density function $p(\bm{x}^0|y)$, characterizing the underlying conditional data distribution, where $y$ represents regression labels. In addressing concerns regarding the potential data insufficiency at $y$, Ding et al.~\cite{ding2021ccgan, ding2023ccgan} posited an assumption wherein minor perturbations to $y$ result in negligible alterations to $p(\bm{x}^0|y)$. Building upon this assumption, they formulated the \textit{Hard and Soft Vicinal Discriminator Losses} (HVDL and SVDL) alongside a novel generator loss, aimed at enhancing the stability of GAN training. Furthermore, to overcome the challenge of encoding regression labels, Ding et al. \cite{ding2021ccgan, ding2023ccgan} proposed both a \textit{Na\"ive Label Input} (NLI) mechanism and an \textit{Improved Label Input} (ILI) mechanism to effectively integrate $y$ into CCGM models. The ILI approach employs a label embedding network $\phi$ comprising a 5-layer perceptron to map the scalar $y$ to a $128\times 1$ vector $\bm{h}^s_y$, i.e.,$\bm{h}^s_y=\phi(y)$. This vector, $\bm{h}^s_y$, is then fed into the generator and discriminator networks of CcGANs to control the generative modeling process. The efficacy of CcGANs has been demonstrated by Ding et al.~\cite{ding2021ccgan, ding2023ccgan} and other researchers~\cite{heyrani2021pcdgan, giry2022sar, zhu2023data} across diverse datasets, showcasing their robustness and versatility in various applications. However, as pointed out in \cite{ding2024turning}, the adversarial training mechanism is vulnerable to data insufficiency, and CcGANs may still generate subpar fake images. Therefore, in this paper, we propose to transition away from the conventional adversarial learning paradigm and instead adopt conditional diffusion processes for CCGM, meanwhile improving the vicinal training and label input mechanisms.
	
	\subsection{Diffusion Models}
	
	As an alternative to GANs, diffusion models offer a viable approach for generative modeling and are widely used in computer vision tasks~\cite{croitoru2023diffusion, 10480591, 10261222, 10321681}. A notable example is \textit{Denoising Diffusion Probabilistic Models} (DDPMs)~\cite{ho2020denoising}, which consist of forward and reverse diffusion processes modeled as Markov chains. The forward diffusion process gradually transforms a real image $\bm{x}^0\sim p(\bm{x}^0)$ into pure Gaussian noise $\bm{x}^T\sim \distNorm(\bm{0},\bm{I})$ through $T$ steps, defined by transitions $q(\bm{x}^t|\bm{x}^{t-1})$, where $T$ is sufficiently large. The reverse process, parameterized by $\bm{\theta}$, reconstructs $\bm{x}^0$ from $\bm{x}^T$ through transitions $p_{\bm{\theta}}(\bm{x}^{t-1}|\bm{x}^t)$. DDPMs are trained using a noise prediction (aka $\bm{\epsilon}$-prediction) loss defined as follows:
	\[
	\mathcal{L}_{simple}(\bm{\theta})=\ex_{\bm{\epsilon},\bm{x}^t}\|\hat{\bm{\epsilon}}_{\bm{\theta}}(\bm{x}^t,t)-\bm{\epsilon}\|^2_2	
	\label{eq:ddpm_loss}
	\] 
	where $\hat{\bm{\epsilon}}_{\bm{\theta}}(\bm{x}^t,t)$ is modeled by a U-Net \cite{ronneberger2015u}. This U-Net is trained to predict the ground truth sampled Gaussian $\bm{\epsilon}\sim \distNorm(\bm{0},\bm{I})$. Although DDPMs can generate high-quality data, they suffer from long sampling times. To mitigate this, Song et al. proposed \textit{Denoising Diffusion Implicit Models} (DDIMs) \cite{song2021denoising}, which introduce a deterministic yet shorter reverse process for high-quality sample generation with significantly fewer steps.
	
	However, DDIM still typically requires hundreds of steps for sample generation, which remains time-consuming. To enable few-step or even one-step sampling, distillation-based methods are employed, with \textit{Improved Distribution Matching Distillation} (DMD2) \cite{yin2024improved} representing a state-of-the-art approach. DMD2 comprises a generator, a real score function, a fake score function, and a discriminator, all modeled with convolutional neural networks. The real score function is a fixed pre-trained U-Net from the diffusion model, while the generator and fake score function, also based on this U-Net, are trainable. The discriminator's feature extraction network shares weights with the encoder of the fake score function and is followed by fully connected layers that output the probability of an input image being real. The training process for DMD2 alternates between two steps: first, optimizing the generator using a distribution matching loss combined with the vanilla GAN loss~\cite{goodfellow2014generative}; second, training the fake score function and the discriminator with an image denoising loss and the vanilla GAN loss. Once training is complete, the generator is used for image generation. Notably, DMD2 multiplies the GAN losses for both the discriminator and generator by two positive weights, $w_D$ and $w_G$, which are treated as hyperparameters.
	
	Furthermore, unlike DDPMs, which aim to estimate the marginal data distribution $p(\bm{x}^0)$, Ho et al. \cite{ho2021classifier} introduced the \textit{Classifier-Free Guidance} (CFG) model to estimate the conditional distribution $p(\bm{x}^0|y)$. As a CDM, CFG integrates class labels as conditions and employs the following equation for sampling~\cite{luo2022understanding}:
	\[
	\triangledown\log p(\bm{x}^t|y)= (1-\gamma) \triangledown\log p(\bm{x}^t) + \gamma \triangledown\log p(\bm{x}^t|y)
	\label{eq:cfg_score_sampling}
	\]
	where $\triangledown\log p(\bm{x}^t|y)$ is an abbreviation of the score function $\triangledown_{\bm{x}^t}\log p(\bm{x}^t|y)$, $y$ represents class labels, and $\gamma$ is a positive hyper-parameter controlling the model's consideration of conditioning information during the sampling procedure.

	\subsection{Text-to-Image Generation} \label{sec:text2img}
	
	\revision{Text-to-image (T2I) generation similarly involves conditional distribution estimation, driving advances in modern diffusion models \cite{rombach2022high, peebles2023scalable, esser2024scaling, labs2025flux1kontextflowmatching}. However, as noted in \cite{ding2023ccgan, ding2024turning} and demonstrated in our experimental results (Section~\ref{sec:experiment}), current T2I architectures like Stable Diffusion \cite{rombach2022high} exhibit critical limitations in CCGM tasks. First, despite their ability to produce realistic images from prompts, these models inherently lack the architectural capacity to address the challenge of scarce or even zero training data for specific regression labels. Second, their dependence on large text encoders like CLIP \cite{radford2021learning} for prompt embedding—significantly more complex than CcGAN's compact 5-layer perceptron $\phi(y)$—introduces substantial and unnecessary memory overhead in CCGM applications.}
	
	Recent work on text-to-image alignment \cite{qu2024discriminative, li2024g} has explored quantitative prompt conditioning, but primarily for discrete counting tasks. While making domain-specific progress, these approaches remain inadequate for CCGM requirements. A key limitation is their inability to properly model $p(\bm{x}|y)$ with $y$ representing regression labels or text containing such labels. Current frameworks typically process only discrete numerical values (e.g., ``Two girls sitting on the grass") rather than continuous regression labels like angular measurements. More critically, these methods inherit the same fundamental limitations as general T2I diffusion models when applied to CCGM tasks, failing to overcome the core challenges of conditional generation in this domain.
	
	\revision{Consequently, quantitative prompt-conditioned image generation remains a largely unexplored area of research.}

	\section{Methodology} \label{sec:methodology}  %
	
\begin{figure}[!htbp]  
		\centering
		\includegraphics[width=1\linewidth]{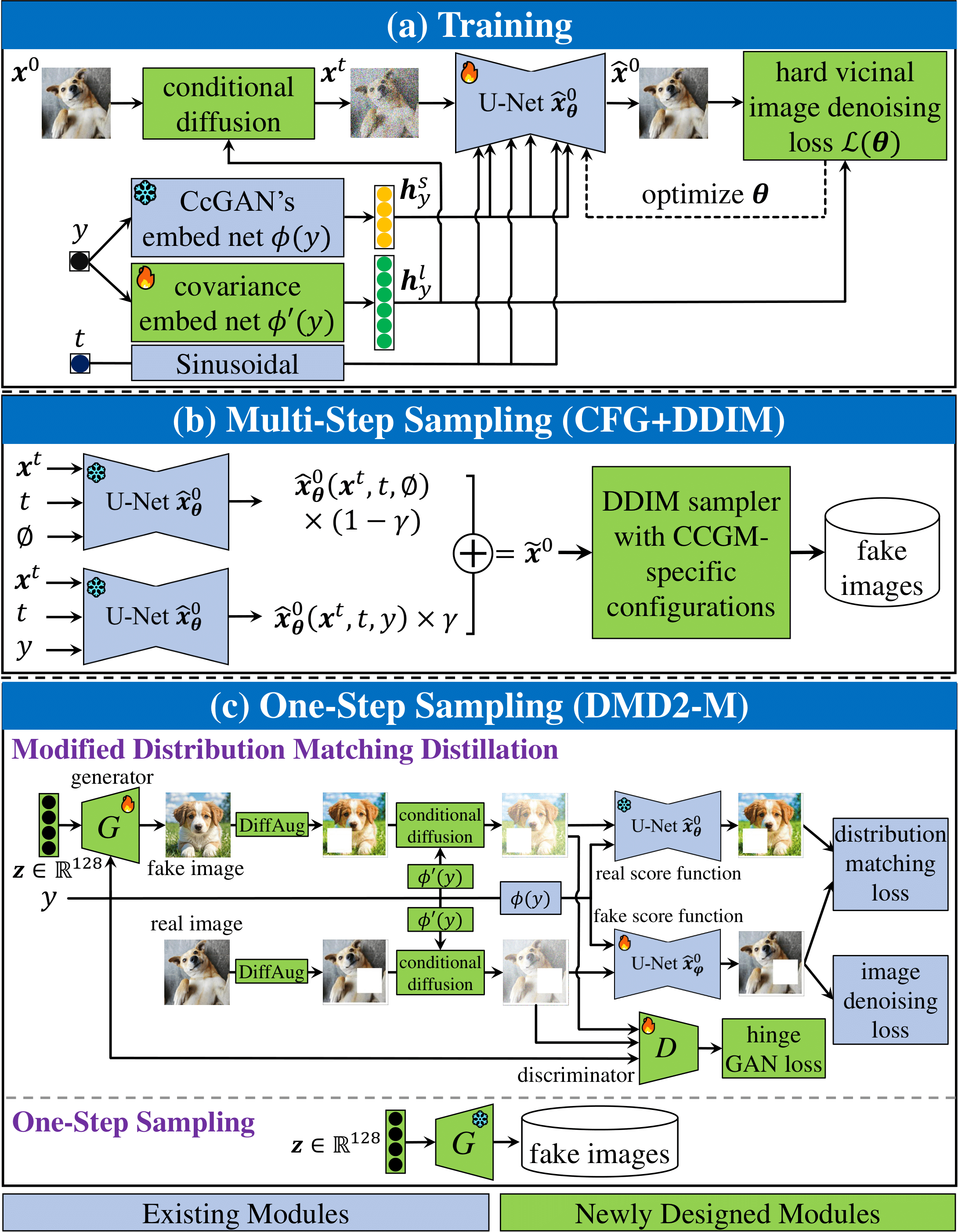}
		\caption{
			\revision{\textbf{The overall workflow of CCDM.} }
			Our method introduces $y$-dependent conditional diffusion processes and a novel hard vicinal image denoising loss (HVIDL) for U-Net training. To integrate the regression label $y$ into CCDM, we use two types of embeddings: a short embedding $\bm{h}_y^s$, encoded by CcGAN’s embedding network for U-Net injection, and a long embedding $\bm{h}_y^l$, encoded by the proposed covariance embedding network for forward diffusion and HVIDL. \revision{For inference, we employ a multi-step sampling algorithm combining CFG~\cite{ho2021classifier} and DDIM~\cite{song2021denoising}, specially optimized for CCGM.} We further introduce DMD2-M, an adaptation of DMD2~\cite{yin2024improved} for one-step CCGM sampling with minimal quality trade-off.
		}
		\label{fig:overall_framework}
	\end{figure}

	In this section, we introduce CCDM, a novel approach specially designed to estimate the conditional distribution $p(\bm{x}^0|y)$ based on $N$ image-label pairs $\{ (\bm{x}^0_i, y_i) \}_{i=1}^N$. CCDM introduces a new $y$-dependent diffusion processes, proposes a novel hard vicinal image denoising loss for training, adapts the CcGAN's label input mechanism to diffusion models, and proposes two efficient sampling algorithms. \revision{The overall workflow is depicted in Fig.~\ref{fig:overall_framework}.}
	
	\subsection{Conditional Diffusion Processes}\label{sec:conditional_diffusion_process}
	
	The proposed CCDM is a conditional diffusion model that comprises two key components: a \textit{forward diffusion process} and a \textit{reverse diffusion process}. In both processes, the regression label $y$ is explicitly incorporated as a conditioning factor.
	
	\subsubsection{Forward Diffusion Process}\label{sec:forward_diffusion}
	
	Given a random sample $\bm{x}^0$ drawn from the actual conditional distribution $\bm{x}^0 \sim p(\bm{x}^0|y)$, we define a forward diffusion process by incrementally adding a small amount of $y$-dependent Gaussian noise over $T$ steps. This process generates a Markov chain comprising a sequence of noisy samples $\bm{x}^1, \ldots, \bm{x}^T$. All noisy samples in this sequence retain the same regression labels $y$ and dimensionality as $\bm{x}^0$. The state transition for this Markov chain is defined as the following $y$-dependent Gaussian model:
	\[
	\label{eq:forward_transition}
	q(\bm{x}^t|\bm{x}^{t-1},y)=\distNorm(\bm{x}^t; \sqrt{1-\beta_t}\bm{x}^{t-1}, \beta_t \bm{H}_y)
	\]
	where $\bm{H}_y=\text{diag}(\exp(-\bm{h}^l_y))$ is a diagonal covariance matrix, and $\bm{h}^l_y$ is a high-dimensional embedding of label $y$ with the same length as flattened $\bm{x}^0$. We train a \textbf{covariance embedding network}, denoted as $\phi^\prime$, comprising five fully-connected layers, to transform the scalar $y$ to the vector $\bm{h}^l_y$, expressed as $\bm{h}^l_y=\phi^\prime(y)$. Further details of $\phi^\prime$ can be found in Section~\ref{sec:label_embed} and Appendix~\ref{sec:supp_cov_label_embedding}. Additionally, the parameters $0\leq\beta_1<...<\beta_T\leq 1$ are fixed to constants using the cosine schedule \cite{nichol2021improved}. Let $\alpha_t=1-\beta_t$ and $\bar{\alpha}_t=\prod_{i=1}^t\alpha_i$, we can sample $\bm{x}^t$ at any arbitrary time step $t$ from
	\[
	\label{eq:q_xt_x0}
	q(\bm{x}^t|\bm{x}^0,y)=\distNorm(\bm{x}^t; \sqrt{\bar{\alpha}_t}\bm{x}^0, (1-\bar{\alpha}_t) \bm{H}_y)
	\]
	Therefore, we have
	\[
	\bm{x}^t = \sqrt{\bar{\alpha}_t}\bm{x}^0+\sqrt{1-\bar{\alpha}_t}\bm{\eps}  \quad\text{where}\quad \bm{\eps}\sim\distNorm(\bm{0}, \bm{H}_y) \label{eq:x_t_from_x0}
	\]

	\subsubsection{Reverse Diffusion Process}\label{sec:reverse_diffusion}
	
	{
		
		In the reverse diffusion process, we aim to leverage the condition $y$ to guide the reconstruction of $\bm{x}^0$ from the Gaussian noise $\bm{x}^T\sim\distNorm(\bm{0}, \bm{H}_y)$. Unfortunately, estimating $q(\bm{x}^{t-1}|\bm{x}^t, y)$ is usually nontrivial. Nevertheless, this reversed conditional distribution is tractable when conditioned on $\bm{x}^0$:
		\[
		\label{eq:q_posterior}
		q(\bm{x}^{t-1}|\bm{x}^t,\bm{x}^0,y)=\distNorm(\bm{x}^{t-1}; \bm{\mu}_q(\bm{x}^t,\bm{x}^0,y), \bm{\Sigma}_q(t,y)).
		\]
		The conditional distribution above is known as the \textbf{ground truth denoising transition}. Specifically,
		\[
		&\bm{\mu}_q(\bm{x}^t,\bm{x}^0,y)= \frac{\sqrt{\alpha_t}(1-\bar{\alpha}_{t-1})\bm{x}^t+\sqrt{\bar{\alpha}_{t-1}}(1-\alpha_t)\bm{x}^0}{1-\bar{\alpha}_t} \label{eq:q_posterior_mean}\\
		&\bm{\Sigma}_q(t,y)=\frac{(1-\alpha_t)(1-\bar{\alpha}_{t-1})}{1-\bar{\alpha}_t}\bm{H}_y \triangleq \sigma_q^2(t)\bm{H}_y. \label{eq:q_posterior_var}
		\]
		The derivation of Eq.~\eqref{eq:q_posterior} can be found in Appendix~\ref{sec:supp_derivation_reverse_process}.
		
		When generating new data points, $\bm{x}^0$ is inherently unknown. Therefore, we train a parametric model to approximate $q(\bm{x}^{t-1}|\bm{x}^t,\bm{x}^0,y)$ using the following form:
		\[
		&p_{\bm{\theta}}(\bm{x}^{t-1}|\bm{x}^t,y)=\distNorm(\bm{x}^{t-1}; \bm{\mu}_{\bm{\theta}}(\bm{x}^t,t,y), \bm{\Sigma}_q(t,y)) \label{eq:p_theta}
		\]
		where,
		\[
		&\bm{\mu}_{\bm{\theta}}(\bm{x}^t,t,y) \\
		= &  \frac{\sqrt{\alpha_t}(1-\bar{\alpha}_{t-1})\bm{x}^t+\sqrt{\bar{\alpha}_{t-1}(1-\alpha_t)}\hat{\bm{x}}_{\bm{\theta}}^0(\bm{x}^t,t,y)}{1-\bar{\alpha}_t} \label{eq:p_theta_mean}
		\]
		In Eq.~\eqref{eq:p_theta_mean}, $\hat{\bm{x}}_{\bm{\theta}}^0(\bm{x}^t,t,y)$ is parameterized by a denoising U-Net~\cite{ronneberger2015u, ho2020denoising}, designed to predict $\bm{x}^0$ based on the noisy image $\bm{x}^t$, timestamp $t$, and condition $y$ associated with both $\bm{x}^t$ and $\bm{x}^0$, where $\bm{\theta}$ denotes the trainable parameters. Further details on this U-Net are provided in Appendix \ref{sec:supp_unet_arch}.
		
	}
	
	\begin{remark}
		For practical demonstration, we opt to use a DDPM-based formulation of diffusion processes, despite the existence of various alternative explanations of diffusion models, such as those grounded in \textit{Stochastic Differential Equations} (SDEs) \cite{song2019generative,song2021scorebased}. Furthermore, in contrast to DDPM, the diffusion processes presented above explicitly integrate conditional information $y$ through the $y$-dependent covariance matrix $\bm{H}_y$.
	\end{remark}
	
	\subsection{Covariance Embedding Network}\label{sec:label_embed}
	
	Although the denoising U-Net for DDPM~\cite{ronneberger2015u, ho2020denoising} supports the sinusoidal positional embedding \cite{ho2020denoising} and Gaussian Fourier projection \cite{song2021scorebased} for encoding time steps $t$, Table~\ref{tab:ab_effect_embedding} shows that these methods are ineffective for embedding regression labels $y$. As illustrated in Fig.~\ref{fig:overall_framework}, $y$ comprises two types of embeddings: \revision{(1) a short embedding $\bm{h}_y^s\in\mathbb{R}^{128}$, which serves as input to the denoising U-Net, and (2) a long embedding $\bm{h}_y^l\in\mathbb{R}^{(C\times W\times H)}$, which modulates the diffusion processes and contributes to both CCDM’s training loss and sampling algorithms, where $C$, $W$, and $H$ denote the number of channels, width, and height of the input images, respectively.} We utilize CcGAN's embedding network $\phi(y)$~\cite{ding2023ccgan} to convert the scalar $y$ into a 128-dimensional vector $\bm{h}^s_y$. However, $\phi(y)$ is unsuitable for computing $\bm{h}_y^l$ due to the significantly higher dimension of $\bm{h}_y^l$ compared to $\bm{h}_y^s$. To address this, we adapt CcGAN's label embedding network to create the \textbf{covariance embedding network}, denoted as $\phi^\prime(y)$, to convert $y$ into $\bm{h}_y^l$. We employ a training scheme nearly identical to \cite{ding2023ccgan} to train $\phi^\prime(y)$, with modifications to the architecture and a reduction in the training epochs of an auxiliary ResNet34 network to suit our requirements. For a more detailed explanation, please refer to Appendix \ref{sec:supp_cov_label_embedding}.
	
	\subsection{Hard Vicinal Image Denoising Loss}\label{sec:model_fitting}
	
	The \textit{Negative Log-Likelihood} (NLL) of CCDM along with its upper bound are defined as follows
	\[
	&-\log\ex_{y\sim p(y)}\left[ \prod_{k=1}^{N^y} f_{\bm{\theta}}(\bm{x}_{k,y}^0|y) \right] \label{eq:cond_loglik} \\
	\leq & \ex_{y\sim p(y)} \left[ \sum_{i=1}^{N} \ind_{\{y_i=y\}} \left( -\log f_{\bm{\theta}}(\bm{x}_{i}^0|y) \right) \right] \label{eq:loglik_lb}
	\]
	where $f_{\bm{\theta}}(\bm{x}^0|y)$ is the likelihood function for observed images with label $y$, $\bm{x}_{k,y}^0$ is the $k$-th image with label $y$, $N^y$ stands for the number of images with label $y$, $p(y)$ is the label distribution, and $\ind$ is the indicator function. Note that Eq.~\eqref{eq:loglik_lb} suggests that estimating $p(\bm{x}^0|y)$ relies solely on images with label $y$, potentially suffering from the data insufficiency issue. 
	
	To address data sparsity, we incorporate the vicinal loss from CcGANs into the training of conditional diffusion models. Assume we use training images with labels in a hard vicinity of $y$ to estimate $p(\bm{x}^0|y)$, then we adjust Eq.~\eqref{eq:loglik_lb} to derive the \textit{\textbf{Hard Vicinal NLL}} (HV-NLL) as follows:
	\[
	& \ex_{y\sim p(y)} \left[ \sum_{i=1}^{N} \ind_{\{|y_i-y|\leq\kappa\}} \left( -\log f_{\bm{\theta}}(\bm{x}_{i}^0|y) \right) \right] \\
	\approx & \frac{C}{N} \sum_{i=1}^N\sum_{j=1}^N \left\{ \ex_{\delta\sim \distNorm(0,\sigma_\delta^2)}\left[ W_h \cdot ( -\log f_{\bm{\theta}}(\bm{x}_i^0|y_j+\delta) ) \right]\right\} \label{eq:vicinal_nll} 
	\]
	where $\sigma_\delta$ is the bandwidth for \textit{Kernel Density Estimation} (KDE) of $p(y)$, $C$ is a constant, and
	\[
	W_h\triangleq\ind_{\{|y_j+\delta-y_i|\leq\kappa\}}.
	\label{eq:W_h}
	\]
	Although not recommended, the hard vicinal weight $W_h$ can be replaced with a soft vicinal weight as follows:
	\[
	W_s\triangleq e^{-\nu(y_j+\delta-y_i)^2}.
	\label{eq:W_s}
	\]
	
	Eq.~\eqref{eq:vicinal_nll} is not analytically tractable, so we minimize its upper bound instead of directly optimizing it. After derivations and simplifications (see Appendix~\ref{sec:supp_detailed_derivation_loss}), we obtain the \textit{\textbf{Hard Vicinal Image Denoising Loss}} (HVIDL) for training the U-Net:
	\[
	\mathcal{L}(\bm{\theta}) = & -\frac{1}{N} \sum_{i=1}^N\sum_{j=1}^N \left\{ \ex_{\delta\sim \distNorm(0,\sigma_\delta^2), \bm{\epsilon}\sim\distNorm(\bm{0},\bm{H}_{y_j+\delta}), t\sim U(1,T)}  \vphantom{\bm{x}^0_i} \right. \\
	&\left. \vphantom{\ex_{\delta\sim \distNorm(0,\sigma_\delta^2)}} \left[ W_h \cdot (\hat{\bm{x}}_{\bm{\theta}}^0 -\bm{x}^0_i)^\intercal \bm{H}_{y_j+\delta}^{-1} (\hat{\bm{x}}_{\bm{\theta}}^0 -\bm{x}^0_i) \right] \right\}. 
	\label{eq:final_loss}
	\]
	Eq.\eqref{eq:final_loss} represents an image denoising (aka $\bm{x}^0$-prediction) loss that is weighted by both the hard vicinal weight $W_h$ and the inverse of the $y$-dependent covariance matrix $\bm{H}_{y_j+\delta}^{-1}$. We minimize this loss for model fitting, with the detailed training procedure outlined in Algorithm \ref{alg:training}.
	
	\begin{remark}
		\label{rmk:vic_param_sel}
		The positive hyperparameters $\kappa$, $\nu$, and $\sigma_\delta$ can be determined using the rule of thumb in Remark 3 of \cite{ding2023ccgan}.
	\end{remark}
	
	\begin{remark}
		\label{rmk:label_drop}
		Our conditional sampling technique, introduced in Section~\ref{sec:conditional_sampling}, builds on classifier-free guidance \cite{ho2021classifier}. Instead of training separate conditional and unconditional diffusion models, we unify them by randomly converting the noisy label $y_j+\delta$ in Eq.~\eqref{eq:final_loss} to $\oslash$ during training, with a \textbf{condition drop probability} $p_{\text{drop}}$. When this happens, the covariance $\bm{H}_y$ becomes the identity matrix $\bm{I}$, reducing the conditional model $\hat{\bm{x}}_{\bm{\theta}}^0(\bm{x}^t,t,y)$ to an unconditional one $\hat{\bm{x}}_{\bm{\theta}}^0(\bm{x}^t,t,\oslash)$.
	\end{remark}
	
	\begin{remark}
		\label{rmk:hard_or_soft}
		The ablation study in Section~\ref{sec:experiment} shows that the hard vicinal weight $W_h$ outperforms the soft one $W_s$. Thus, we recommend using the hard vicinal weight $W_h$ in applications.
	\end{remark}
	
	\begin{remark}
		\label{rmk:pred_x0_or_noise}
		It can be shown that Eq.~\eqref{eq:final_loss} can be transformed into a $\bm{\epsilon}$-prediction loss by utilizing the relationship between $\bm{x}^0$-prediction and $\bm{\epsilon}$-prediction \cite{ho2020denoising, luo2022understanding}. However, when using the accelerated sampler DDIM~\cite{song2021denoising}, this $\bm{\epsilon}$-prediction loss often leads to significant label inconsistency, as evidenced in Table~\ref{tab:ab_effect_objective}.
	\end{remark}

	\subsection{CFG and DDIM-based Multi-Step Sampling}\label{sec:conditional_sampling}
	
	To generate samples from the learned model, we adapt the CFG framework~\cite{dhariwal2021diffusion}, originally designed for class-conditional sampling, to CCGM. Given a regression label $y$, we aim to generate corresponding fake images. Luo~\cite{luo2022understanding} demonstrated a linear relationship between $\bm{x}^0$ and $\triangledown\log p(\bm{x}^t|y)$ as follows:
	\[
	\bm{x}^0 = \frac{\bm{x}^t+(1-\bar{\alpha}_t)\triangledown\log p(\bm{x}^t|y)}{\sqrt{\bar{\alpha}_t}}
	\label{eq:linear_x0_score}
	\]
	We then modify Eq. \eqref{eq:cfg_score_sampling} to reconstruct $\bm{x}^0$:
	\[
	\tilde{\bm{x}}^0 = (1-\gamma)\cdot \underbrace{\hat{\bm{x}}_{\bm{\theta}}^0(\bm{x}^t,t,\oslash)}_{\text{unconditional}} + \gamma \cdot \underbrace{\hat{\bm{x}}_{\bm{\theta}}^0(\bm{x}^t,t,y)}_{\text{conditional}},
	\label{eq:cfg_x0_sampling}
	\]
	where \revision{$\gamma$ (referred to as the \textbf{conditional scale}) is recommended as $\gamma \in [1.5, 2]$, alongside $p_{\text{drop}} = 0.1$, to balance image diversity and label consistency in the CCGM task. This contrasts with class-conditional or text-to-image scenarios, where typical settings are $\gamma > 3$ and $p_{\text{drop}} \geq 0.2$ \cite{dhariwal2021diffusion, balaji2022ediff}.}
	
	For fast sampling, we use the DDIM sampler \cite{song2021denoising} to generate new samples with $T^\prime\ll T$ time steps through the following deterministic procedure:
	\[
	\bm{x}^{t-1} = \sqrt{\bar{\alpha}_{t-1}}\tilde{\bm{x}}^0 + \sqrt{1-\bar{\alpha}_{t-1}}\frac{\bm{x}^t-\sqrt{\bar{\alpha}_{t}}\tilde{\bm{x}}^0}{\sqrt{1-\bar{\alpha}_{t}}},
	\]
	where \revision{the initial $\bm{x}^{T^\prime}$ is randomly sampled from  $\distNorm(\bm{0},\bm{H}_y)$ instead of a standard Gaussian $\distNorm(\bm{0},\bm{I})$, making the sampling algorithm (Algorithm~\ref{alg:multistep_sampling}) different from \cite{song2021denoising}.}

	\subsection{Distillation-based One-Step Sampling}\label{sec:dmd_sampling}
	
\begin{figure}[!htbp] 
		\centering
		\includegraphics[width=0.7\linewidth]{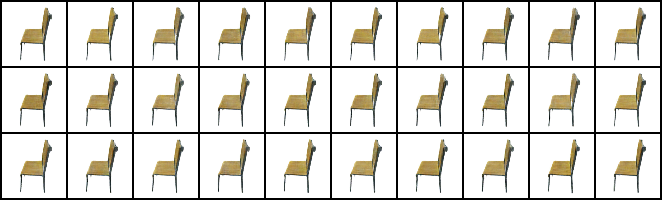}
		\caption{DMD2 \cite{yin2024improved} suffers from severe mode collapse on RC-49 ($64 \times 64$), demonstrating its inapplicability to the CCGM task.}
		\label{fig:vanilla_dmd2_failure_example}
	\end{figure}

	Even with DDIM, conditional sampling with CFG remains time-consuming, often taking hours or days to generate sufficient evaluation samples. To enhance efficiency, we attempted to distill the trained CCDM into a one-step generator using DMD2~\cite{yin2024improved}, but its model design and training approach proved ineffective for our CCGM task. For instance, when applied to RC-49 ($64\times 64$), we observed severe mode collapse, as shown in Fig.~\ref{fig:vanilla_dmd2_failure_example} and Table~\ref{tab:ablation_dmd2m}. Furthermore, one-step sampling does not require a U-Net architecture, and sharing the feature extraction head between the discriminator and fake score function harms the discriminator's performance, as these models need distinct feature representations. Additionally, DMD2's reliance on vanilla GAN loss and outdated training techniques makes it unsuitable for CCGM tasks, which often face data insufficiency and imbalance. To address these issues, we propose DMD2-M, an enhanced version of DMD2 with the following improvements:
	
	\begin{itemize}
		
		\item \textbf{Network architecture:}  DMD2-M adopts the SNGAN~\cite{miyato2018spectral} architecture for both generator $G$ and discriminator $D$, using a 128-dimensional Gaussian noise for the generator instead of a high-dimensional noisy image. The discriminator is trained independently and does not share a feature extraction head with the fake score function model.
		
		\item \textbf{Loss function:} DMD2-M significantly improves upon DMD2 by adopting the hinge GAN loss \cite{miyato2018spectral} in place of the vanilla GAN loss. Moreover, it incorporates the hard vicinal weight (Eq. \eqref{eq:W_h}) into all training losses to help mitigate mode collapse in data-limited scenarios. Additionally, our proposed $y$-dependent covariance matrix $\bm{H}_y$ is integrated into the forward diffusion process for the computation of training losses.
		
		\item \textbf{Training technique:} DMD2-M integrates DiffAugment~\cite{zhao2020differentiable} into the distillation process—a highly effective and user-friendly technique that stabilizes adversarial training and improves performance, especially in situations with limited training data. 
		
	\end{itemize}
	
	\begin{remark}
		By default, CCDM utilizes DDIM as its sampler to ensure optimal image quality unless stated otherwise. In contrast, DMD2-M is tailored for scenarios demanding rapid sampling, where a slight compromise in quality and extended training time are acceptable.
	\end{remark}
	
	\begin{algorithm}[!htbp] 
		\caption{Algorithm for Training CCDM}
		\label{alg:training}  
		\KwData{$N$ real image-label pairs $\Omega=\{(\bm{x}^0_i, y_i)\}_{i=1}^N$, $\tilde{N}$ distinct labels $\Upsilon=\{y_{[i]}\}_{i=1}^{\tilde{N}}$ arranged in ascending order from the training dataset;}
		\KwIn{The pre-trained CcGAN's embedding network $\phi(y)$ \cite{ding2023ccgan}; The pre-trained covariance embedding network $\phi^\prime(y)$; Preset hard vicinity radius $\kappa$, KDE bandwidth $\sigma_\delta$, time steps $T$, condition drop probability $p_\text{drop}$, number of iterations $K$, and batch size $m$;}
		\KwResult{A trained U-Net $\hat{\bm{x}}_{\bm{\theta}}^0$ }
		\For{$k=1$ \KwTo $K$}{
			Draw $m$ regression labels, denoted by $Y$, with replacement from $\Upsilon$\;
			Create a batch of target labels $Y^{\delta}=\{ y_i+\delta| y_i\in Y, \delta\sim \mathcal{N}(0,\sigma_\delta^2), i=1,\dots,m \}$ ($\hat{\bm{x}}_{\bm{\theta}}^0$ is conditional on these labels)  \;
			Initialize $\Omega^{y_i+\delta}=\emptyset$ \;
			\For{$i=1$ \KwTo $m$}{
				Randomly choose an image-label pair $(\bm{x}^0,y) \in \Omega$ satisfying $|y-y_i-\delta|\leq\kappa$ where $y_i+\delta\in Y^{\delta}$\; 
				Draw $u\sim U(0,1)$\;
				\uIf{$u\geq p_\text{drop}$}{
					Update $\Omega^{y_i+\delta}=\Omega^{y_i+\delta} \cup (\bm{x}^0, y_i+\delta)$\;
				}
				\Else{
					Update $\Omega^{y_i+\delta}=\Omega^{y_i+\delta} \cup (\bm{x}^0, \emptyset)$\;
				}
			}
			Convert regression labels in $\Omega^{y_i+\delta}$ into their multi-dimensional representations $\bm{h}_{y_i+\delta}^{s}$ and $\bm{h}_{y_i+\delta}^{l}$ by applying the functions $\phi(y)$ and $\phi^\prime(y)$, respectively \; 
			Compute the $y$-dependent covariance matrix $\bm{H}_{y_i+\delta}$ via $\bm{H}_{y_i+\delta}=\text{diag}(\exp(-\bm{h}^l_{y_i+\delta}))$ \;
			Update $\hat{\bm{x}}_{\bm{\theta}}^0$ using samples in the set $\Omega^{y_i+\delta}$ via gradient-based optimizers according to Eq.~\eqref{eq:final_loss}\; 
		}
	\end{algorithm}

	\begin{algorithm}[!htbp] 
		\caption{Multi-Step Sampling Algorithm for CCDM Utilizing CFG and DDIM}
		\label{alg:multistep_sampling}  
		\KwIn{The pre-trained U-Net $\hat{\bm{x}}_{\bm{\theta}}^0$; The pre-trained CcGAN's embedding network $\phi(y)$  \cite{ding2023ccgan}; The pre-trained covariance embedding network $\phi^\prime(y)$; The target regression label $y$; Preset hyperparameters, including the time steps $T^\prime$ for sampling, the conditional scale $\gamma$, and the variance schedule $\beta_i$'s;}
		\KwResult{A fake image $\bm{x}^0_g$ with label $y$}
		Convert the regression label $y$ into its multi-dimensional representations $\bm{h}_{y}^{s}$ and $\bm{h}_{y}^{l}$ by applying the functions $\phi(y)$ and $\phi^\prime(y)$, respectively \; 
		Compute the $y$-dependent covariance matrix $\bm{H}_y$ via $\bm{H}_y=\text{diag}(\exp(-\bm{h}^l_y))$ \;
		$t\leftarrow T^\prime$, $\bm{x}^{T^\prime}_g\sim \distNorm(\bm{0},\bm{H}_y)$ \;
		\While{$t > 0$}{
			Compute $\alpha_t=1-\beta_t$ and $\bar{\alpha}_t=\prod_{i=1}^t\alpha_i$\;
			Compute $\tilde{\bm{x}}^0 = (1-\gamma)\cdot \hat{\bm{x}}_{\bm{\theta}}^0(\bm{x}^t_g,t,\oslash) + \gamma \cdot \hat{\bm{x}}_{\bm{\theta}}^0(\bm{x}^t_g,t,y) $\;
			Update $\bm{x}^{t-1}_g = \sqrt{\bar{\alpha}_{t-1}}\tilde{\bm{x}}^0 + \sqrt{1-\bar{\alpha}_{t-1}}\frac{\bm{x}^t_g-\sqrt{\bar{\alpha}_{t}}\tilde{\bm{x}}^0}{\sqrt{1-\bar{\alpha}_{t}}}$\;
			Update $t\leftarrow t-1$\;}
	\end{algorithm}

	\section{Experiments}\label{sec:experiment}
	
	\subsection{Experimental Setup}\label{sec:experimental_setup}
	
	{\setlength{\parindent}{0cm}\textbf{Datasets.}}  We evaluate the effectiveness of CCDM across four datasets with varying resolutions ranging from $64\times 64$ to $192\times 192$. These datasets comprise RC-49~\cite{ding2023ccgan}, UTKFace~\cite{utkface}, Steering Angle~\cite{steeringangle}, and Cell-200~\cite{ding2023ccgan}. Brief introductions to these datasets are presented below:
	\begin{itemize}
		\item \textbf{RC-49}: The RC-49 dataset consists of 44,051 RGB images of 49 chair types at $64 \times 64$ resolution. Each chair type is labeled with 899 images at yaw rotation angles ranging from $0.1^{\circ}$ to $89.9^{\circ}$ in $0.1^{\circ}$ increments. For training, we select angles with odd last digits and randomly sample 25 images per angle, resulting in a training set of 11,250 images corresponding to 450 distinct yaw angles.
		
		\item \textbf{UTKFace}: The UTKFace dataset comprises 14,760 RGB human face images with ages as regression labels, spanning from 1 to 60 years \cite{ding2021ccgan}. It features 50 to 1051 images per age, all used for training. The dataset is available in three resolutions: $64\times 64$, $128\times 128$, and $192\times 192$.
		
		\item \textbf{Steering Angle}: The Steering Angle dataset, derived from an autonomous driving dataset \cite{steeringangle}, is preprocessed by Ding et al. \cite{ding2023ccgan} for our experiments. It consists of 12,271 RGB images captured by a dashboard-mounted camera, available in resolutions of $64 \times 64$ and $128 \times 128$. These images are labeled with 1,774 unique steering wheel rotation angles, ranging from $-80.0^\circ$ to $80.0^\circ$.
		
		\item \textbf{Cell-200}: The Cell-200 dataset, created by Ding et al.~\cite{ding2023ccgan}, contains 200,000 synthetic grayscale microscopic images at a $64 \times 64$ resolution. Each image contains a variable number of cells, ranging from 1 to 200, with 1,000 images available for each cell count. For training, we select a subset with only images containing an odd number of cells, choosing 10 images per cell count, resulting in 1,000 training images. For evaluation, we use all 200,000 samples from the dataset.
		
	\end{itemize}

	{\setlength{\parindent}{0cm}\textbf{Compared Methods.}} In accordance with the setups described in \cite{ding2024turning}, we choose the following contemporary conditional generative models for comparison: 
	\begin{itemize}
		\item \textbf{Two class-conditional GANs}: We compare against ReACGAN~\cite{kang2021rebooting} and ADCGAN~\cite{hou2022conditional}, both leveraging state-of-the-art architectures and training techniques with discretized regression labels.
		
		\item \textbf{Two class-conditional diffusion models}: For comparison, we select two classic CDMs, ADM-G~\cite{dhariwal2021diffusion} and Classifier-Free Guidance (CFG)\cite{ho2021classifier}, both conditioned on discretized regression labels. We exclude more advanced CDMs like Stable Diffusion (SD)\cite{rombach2022high} and DiT~\cite{peebles2023scalable} for the following reasons: This study serves as a proof of concept, focusing on demonstrating how our modifications to diffusion processes, conditioning mechanisms, and the introduction of a new vicinal loss enable effective application of CDMs to CCGM. These enhancements are applicable to state-of-the-art CDMs like SD and DiT but require substantial computational resources beyond our current capabilities. Therefore, we base CCDM on DDPM~\cite{ho2020denoising} and CFG~\cite{ho2021classifier} and ensure a fair comparison by evaluating it against ADM-G and CFG, which have similar model capacities.
		
		\item \textbf{A text-to-image diffusion model}: Our comparative evaluation on RC-49 and UTKFace incorporates Stable Diffusion v1.5 (SD v1.5) \cite{rombach2022high}, utilizing full fine-tuning of the official Hugging Face checkpoint (stable-diffusion-v1-5). We exclude more advanced Stable Diffusion 3 (SD 3) \cite{esser2024scaling} based on preliminary experiments, which reveal that SD 3 Medium demands substantially greater computational resources (both in sampling time and GPU memory) than SD v1.5 while exhibiting inferior performance on RC-49—with particular deficiencies in output realism and label consistency.
		
		\item \textbf{Two GAN-based CCGM models}: We compare against state-of-the-art CcGAN (SVDL+ILI)~\cite{ding2023ccgan} and Dual-NDA~\cite{ding2024turning}, a data augmentation strategy designed to enhance CcGAN's visual quality and label consistency. Both methods are based on the CcGAN framework.
		
		\item \textbf{Two diffusion-based CCGM models}: We include CcDPM~\cite{zhao2024ccdpm} and our proposed CCDM in the comparison. CcDPM, originally designed for non-image data, is adapted for the CCGM task by replacing its U-Net backbone and label embedding mechanism with our proposed approach. However, we retained CcDPM's soft vicinity and $\bm{\epsilon}$-prediction loss.
		
	\end{itemize}
	For the $192\times 192$ experiment, we focus exclusively on SD v1.5, CcGAN (SVDL+ILI), Dual-NDA, CcDPM, and CCDM.

	{\setlength{\parindent}{0cm}\textbf{Implementation Setup.}} 
	In our implementation of class-conditional models, we bin the regression labels of four datasets into various classes: 150 classes for RC-49, 60 classes for UTKFace, 221 classes for Steering Angle, and 100 classes for Cell-200. To facilitate comparisons, we leveraged pre-trained models provided by Ding et al.\cite{ding2024turning} for experiments on UTKFace ($64\times 64$ and $128\times 128$) and Steering Angle ($64\times 64$ and $128\times 128$) datasets, rather than re-implementing all candidate methods (excluding CCDM). Additionally, Ding et al.\cite{ding2023ccgan} offered pre-trained ReACGAN for RC-49 and CcGAN (SVDL+ILI) for UTKFace ($192\times 192$). For other candidate models, we re-implemented them for comparison. When implementing CcGAN (SVDL+ILI) and Dual-NDA on Cell-200, we used DCGAN \cite{radford2015unsupervised} as the backbone and trained with the vanilla cGAN loss \cite{mirza2014conditional}. For RC-49, we employed SAGAN \cite{zhang2019self} as the backbone and trained with hinge loss \cite{lim2017geometric} and DiffAugment \cite{zhao2020differentiable}. For CCDM, we set $p_{\text{drop}}=0.1$ in Remark~\ref{rmk:label_drop} and $\gamma=1.5$ in Eq.~\eqref{eq:cfg_x0_sampling} for most experiments, except for UTKFace ($128\times 128$ and $192\times 192$), where we increased $\gamma$ to 2.0 for better label consistency. The total time step $T$ is fixed at 1000 during training, while the sampling time step $T^\prime$ varies: 250 for the $64\times 64$ experiments, and 150 for all other experiments. The hyperparameters $\kappa$ and $\sigma_\delta^2$ were set according to the rule of thumb in Remark 3 of \cite{ding2023ccgan}. The training configurations of CcDPM were largely similar to those of CCDM, with the exception that CcDPM utilized $\bm{\epsilon}$-prediction loss and employed the soft vicinal weight. The one-step sampler, DMD2-M, is implemented only in the RC-49 ($64\times 64$) and Steering Angle ($64\times 64$) experiments, where it is compared with the DDPM and DDIM samplers. \revision{Stable Diffusion v1.5 (an advanced text-to-image model in our comparison) is fine-tuned on both RC-49 and UTKFace datasets, using the prompt templates ``a photo of a chair at yaw angle [ANGLE] degrees" and ``a portrait of the face of [AGE] year old", respectively. } The detailed implementation configurations are available in our GitHub repository.

	{\setlength{\parindent}{0cm}\textbf{Evaluation Setup.}} Consistent with the work of Ding et al.~\cite{ding2021ccgan, ding2023ccgan, ding2023efficient, ding2024turning}, we predefine $m_c$ distinct points within the regression label range for each experiment, which we refer to as \textbf{evaluation centers}. For each evaluation center, we generate $N_c^g$ fake images, resulting in a total of 179,800, 60,000, 100,000, and 200,000 images generated by each candidate method in the RC-49, UTKFace, Steering Angle, and Cell-200 experiments, respectively. These images are evaluated using one overall metric and three separate metrics. The overall metric is the \textit{Sliding Fr\'echet Inception Distance} (SFID)~\cite{ding2023ccgan}, calculated as the average FID score across evaluation centers, with the standard deviation of these FIDs reported in Table ~\ref{tab:main_results}. Additionally, we use \textit{Naturalness Image Quality Evaluator} (NIQE)~\cite{mittal2012making} for visual fidelity, Diversity~\cite{ding2023ccgan} for image variety, and Label Score~\cite{ding2023ccgan} for label consistency. Note that the Diversity score cannot be computed for the Cell-200 dataset due to the lack of categorical image labels. For SFID, NIQE, and Label Score, lower values indicate better performance, while higher Diversity scores are preferred for greater variety in generated samples. For detailed evaluation setups, please refer to Appendix \ref{sec:supp_detailed_test_setups}.

	\subsection{Experimental Results} 
	
\begin{figure*}[!htbp]
		\centering
		\includegraphics[width=1\linewidth]{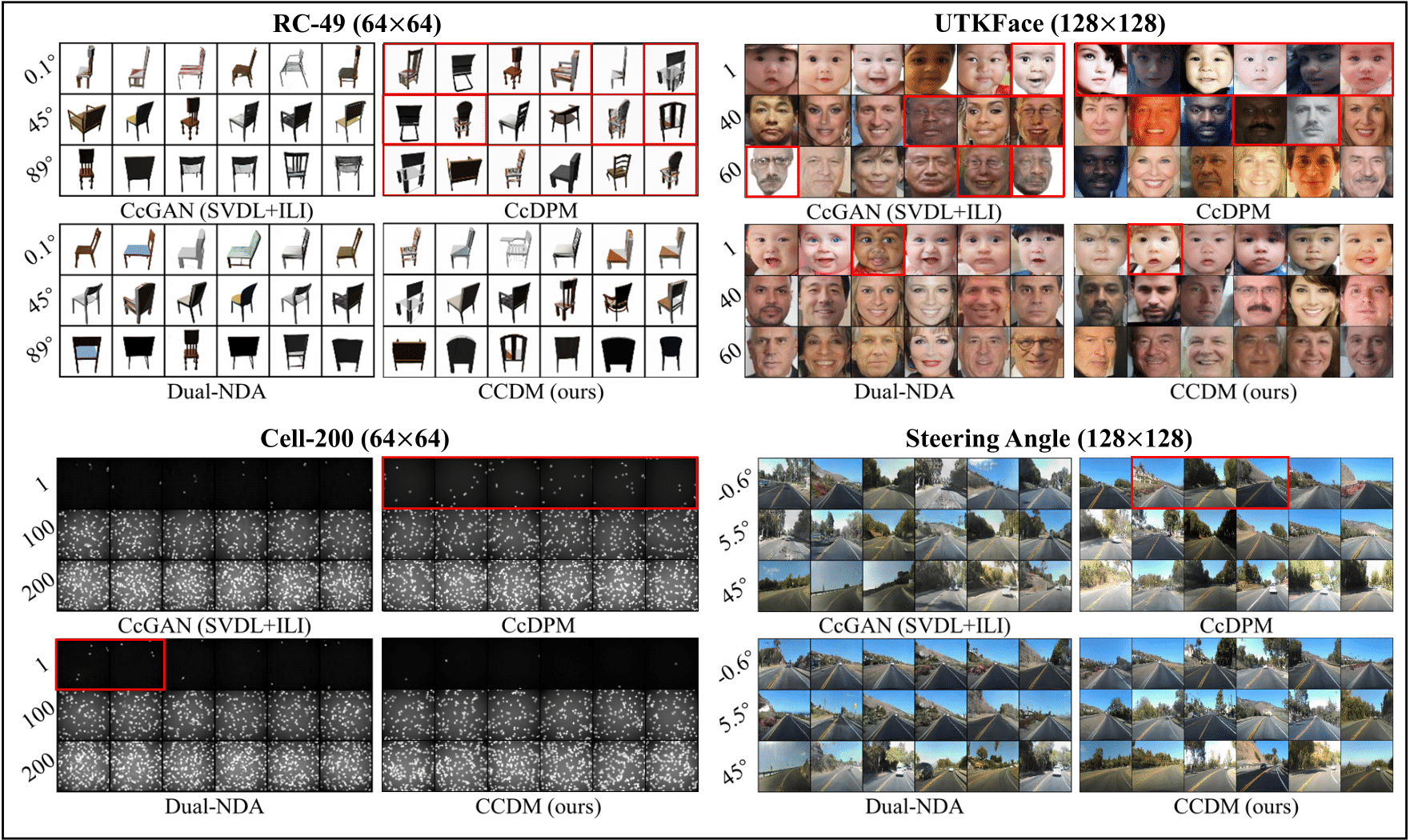}
		\caption{Example fake images generated by four CCGM methods across four datasets, with obvious deficiencies marked in red.}
		\label{fig:example_fake_images}
	\end{figure*}

\begin{figure}[!htbp]
		\centering
		\includegraphics[width=1\linewidth]{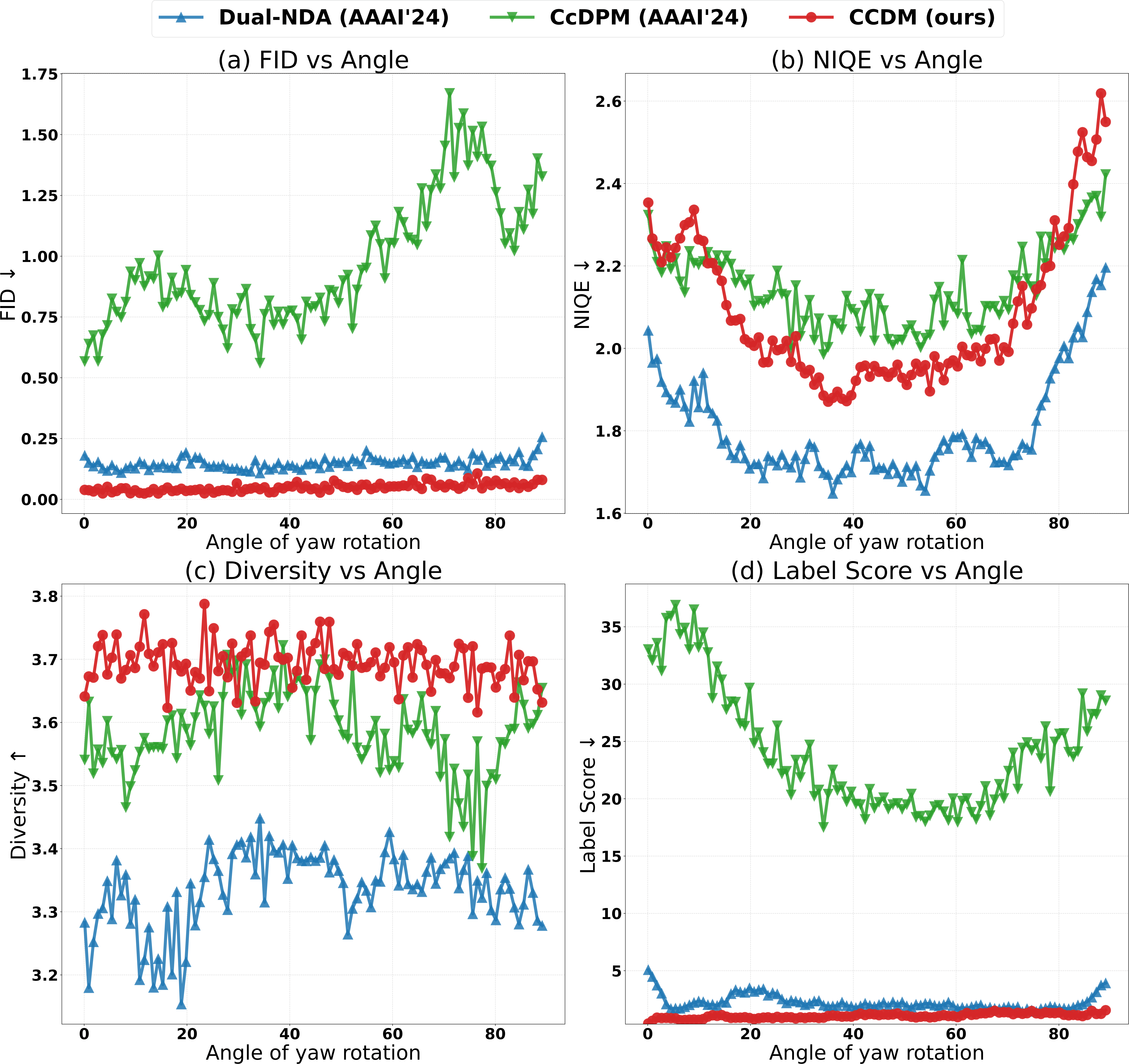} 
		\caption{ \revision{Line graphs of FID/NIQE/Diversity/Label Score versus angle (evaluation center) for three compared methods on RC-49 ($64\times 64$).} } 
		\label{fig:line_graphs_rc49}
	\end{figure}

	
	We evaluate candidate methods across seven experimental settings spanning four datasets and three resolutions, with complete quantitative results documented in Table~\ref{tab:main_results}. Representative visual results are shown in Fig.~\ref{fig:line_graphs_rc49} and Fig.~\ref{fig:example_fake_images}. \revision{Sampling cost comparisons for CCGM methods are performed on RC-49, with quantitative results detailed in Table~\ref{tab:ab_sampling_cost}.} The performance of DMD2-M against three other samplers is summarized in Table~\ref{tab:ab_effect_samplers}. Our analysis of the experimental results highlights the following findings:
	
	\begin{itemize}
		\item \textbf{CCDM achieves the lowest SFID scores across all settings, indicating superior overall performance.} Notably, in the RC-49 ($64\times 64$) experiment, its SFID score is half that of CcGAN and Dual-NDA, and only 5\% of CcDPM's, demonstrating significant image quality improvements. 
		
		\item Consistent with \cite{ding2023ccgan, ding2024turning}, class-conditional models perform significantly worse than CCGM. Notably, class-conditional GANs suffer from mode collapse in the Cell-200 experiment and frequent label inconsistency across settings.
		
		\item Across all datasets except RC-49, CCDM achieves lower NIQE scores than the baseline CcGAN (SVDL+ILI), indicating superior visual fidelity. Even against Dual-NDA, designed to enhance CcGAN’s visual quality, CCDM outperforms in 4 of 7 settings.
		
		\item The Diversity scores vary without a consistent trend. In some settings, CCDM achieves the highest diversity, while in others, it is notably lower yet remains above average.
		
		\item In terms of label consistency, CCDM consistently outperforms or matches the performance of CcGAN and Dual-NDA across all settings.
		
		\item In all settings, CCDM significantly outperforms CcDPM, and CcDPM generally lags behind GAN-based CCGM models. Notably, CcDPM shows poor visual quality and label consistency on RC-49, as evidenced by Fig.~\ref{fig:line_graphs_rc49}, where CCDM outperforms it across all angles for evaluation in the $64\times 64$ experiment.
		
		\item Table~\ref{tab:ab_effect_samplers} shows that DDPM1K provides the best image quality but requires much longer sampling time, while DDIM250 offers the second-best results. Although DMD2-M sacrifices some image quality compared to DDIM250, it reduces sampling time from hours or days to seconds. Considering both Tables~\ref{tab:main_results} and \ref{tab:ab_effect_samplers}, DMD2-M still outperforms CcGAN and Dual-NDA despite the quality trade-off.
		
		\item \revision{While CCDM ($T^\prime=250$) achieves the lowest SFID scores, this comes at the cost of substantially higher computational requirements. Although DMD2-M demonstrates a viable trade-off between image quality and sampling efficiency (reducing both time and memory usage), developing more efficient sampling algorithms for CCDM remains an important direction for future research.}
		
	\end{itemize} 
	

	\begin{table*}[!htbp] 
		\begin{adjustbox}{width=1\textwidth}
			\begin{threeparttable}
				\centering
				\caption{ \revision{Average Quality of Fake Images Generated from Compared Methods} }
				\begin{tabular}{cccccccc}
					\toprule
					\begin{tabular}[c]{@{}c@{}} \textbf{Dataset} \\ (resolution)\end{tabular} & \begin{tabular}[c]{@{}c@{}} \textbf{Condition} \\ \textbf{Type} \end{tabular} & \begin{tabular}[c]{@{}c@{}} \textbf{Framework} \\ \textbf{Type} \end{tabular} & \textbf{Method} & \begin{tabular}[c]{@{}c@{}} \textbf{SFID} $\downarrow$ \\ (overall quality)\end{tabular}   & \begin{tabular}[c]{@{}c@{}} \textbf{NIQE} $\downarrow$ \\ (visual fidelity)\end{tabular} & \begin{tabular}[c]{@{}c@{}} \textbf{Diversity} $\uparrow$ \\ {(diversity)}\end{tabular} & \begin{tabular}[c]{@{}c@{}} \textbf{Label Score} $\downarrow$ \\ (label consistency)\end{tabular} \\
					\midrule
					
					\multirow{8}[2]{*}{\begin{tabular}[c]{@{}c@{}} \textbf{RC-49} \\ {(64$\times$64)}\end{tabular}} & \multirow{4}[0]{*}{\begin{tabular}[c]{@{}c@{}} Categorical\\ {(150 classes)}\end{tabular}} & \multirow{2}[0]{*}{GAN} & ReACGAN (Neurips'21)~\cite{kang2021rebooting} & 0.178 (0.036) & 1.938 (0.141) & 3.145 (0.028) & {1.422 (1.337)} \\
					&    &   & ADCGAN (ICML'22)~\cite{hou2022conditional} & 1.158 (0.171) & 2.364 (0.107) & 2.714 (0.053) & 30.547 (21.635) \\
					\cdashline{3-8}
					\specialrule{0em}{1pt}{1pt}
					&    &\multirow{2}[0]{*}{Diffusion}& ADM-G (Neurips'21)~\cite{dhariwal2021diffusion} & 0.641 (0.221) & 2.504 (0.229) & 2.418 (0.275) & 4.635 (7.386) \\
					&    &   & CFG (Neurips'21)~\cite{ho2021classifier}   & 2.657 (0.753) & 2.850 (0.375) & 2.350 (0.557) & 52.312 (17.857) \\
					\cdashline{2-8}
					\specialrule{0em}{1pt}{1pt}
					
					& Text & Diffusion & SD v1.5 (CVPR'22)~\cite{rombach2022high} & 0.546 (0.126) & 2.910 (0.222) & 1.730 (0.232) & 10.743 (11.187) \\
					\cdashline{2-8}
					\specialrule{0em}{1pt}{1pt}
					
					& \multirow{4}[0]{*}{Continuous} & \multirow{2}[0]{*}{GAN} & CcGAN (T-PAMI'23)~\cite{ding2023ccgan} & {0.126 (0.034)} & {1.809 (0.144)}  & {3.451 (0.055)} & 2.655 (2.162) \\
					&   &    & Dual-NDA (AAAI'24)~\cite{ding2024turning} & 0.148 (0.025) & \textbf{1.808 (0.133)} & 3.344 (0.060) & 2.211 (1.912) \\
					\cdashline{3-8}
					\specialrule{0em}{1pt}{1pt}
					& & \multirow{2}[0]{*}{Diffusion}  & CcDPM (AAAI'24)~\cite{zhao2024ccdpm} & 0.970 (0.254) & 2.153 (0.098) & 3.581 (0.069) & 24.174 (19.689) \\
					&   &    & \textbf{CCDM (ours)} & \textbf{0.049 (0.016)} & 2.086 (0.182) & \textbf{3.698 (0.037)} & \textbf{1.074 (0.978)} \\
					\midrule
					
					\multirow{8}[2]{*}{\begin{tabular}[c]{@{}c@{}} \textbf{UTKFace} \\ {(64$\times$64)}\end{tabular}} & \multirow{4}[0]{*}{\begin{tabular}[c]{@{}c@{}} Categorical\\ {(60 classes)}\end{tabular}} & \multirow{2}[0]{*}{GAN} & ReACGAN (Neurips'21)~\cite{kang2021rebooting} & 0.548 (0.136) & 1.679 (0.170) & 1.206 (0.240) & 6.846 (5.954) \\
					&   &    & ADCGAN (ICML'22)~\cite{hou2022conditional} & 0.573 (0.218) & 1.680 (0.140) & \textbf{1.451 (0.019)} & 17.574 (12.388) \\
					\cdashline{3-8}
					\specialrule{0em}{1pt}{1pt}
					&    & \multirow{2}[0]{*}{Diffusion}   & ADM-G (Neurips'21)~\cite{dhariwal2021diffusion} & 0.744 (0.195) & 2.856 (0.225) & 0.917 (0.318) & 7.583 (6.066) \\
					&    &   & CFG (Neurips'21)~\cite{ho2021classifier} & 2.155 (0.638) & 1.681 (0.303) & 0.858 (0.413) & 8.477 (7.820) \\
					\cdashline{2-8}
					\specialrule{0em}{1pt}{1pt}
					
					& Text & Diffusion & SD v1.5 (CVPR'22) \cite{rombach2022high} & 1.013 (0.271) & 2.607 (0.453) & 1.091 (0.310) & \textbf{5.027 (3.839)} \\
					\cdashline{2-8}
					\specialrule{0em}{1pt}{1pt}
					
					& \multirow{4}[0]{*}{Continuous} & \multirow{2}[0]{*}{GAN} & CcGAN (T-PAMI'23)~\cite{ding2023ccgan} & 0.413 (0.155) & 1.733 (0.189) & {1.329 (0.161)} & 8.240 (6.271) \\
					&    &   & Dual-NDA (AAAI'24)~\cite{ding2024turning} & {0.396 (0.153)} & {1.678 (0.183)} & 1.298 (0.187) & {6.765 (5.600)} \\
					\cdashline{3-8}
					\specialrule{0em}{1pt}{1pt}
					&    & \multirow{2}[0]{*}{Diffusion}  & CcDPM (AAAI'24)~\cite{zhao2024ccdpm} & 0.466 (0.140) & 1.560 (0.143) & 1.211 (0.238) & 6.868 (5.821) \\
					&   &    & \textbf{CCDM (ours)} & \textbf{0.363 (0.134)} & \textbf{1.542 (0.153)} & 1.184 (0.268) & 6.164 (5.173) \\
					\midrule
					
					\multirow{8}[2]{*}{\begin{tabular}[c]{@{}c@{}} \textbf{Steering Angle} \\ {(64$\times$64)}\end{tabular}} & \multirow{4}[0]{*}{\begin{tabular}[c]{@{}c@{}} Categorical\\ {(221 classes)}\end{tabular}} & \multirow{2}[0]{*}{GAN} & ReACGAN (Neurips'21)~\cite{kang2021rebooting} & 3.635 (0.491) & 2.099 (0.072) & 0.543 (0.366) & 27.277 (21.508) \\
					&    &   & ADCGAN (ICML'22)~\cite{hou2022conditional} & 2.960 (1.083) & 2.015 (0.003) & 0.930 (0.018) & 40.535 (24.031) \\
					\cdashline{3-8}
					\specialrule{0em}{1pt}{1pt}
					&    &  \multirow{2}[0]{*}{Diffusion} & ADM-G (Neurips'21)~\cite{dhariwal2021diffusion} & 2.890 (0.547) & 2.164 (0.200) & 0.205 (0.160) & 24.186 (20.685) \\
					&    &   & CFG (Neurips'21)~\cite{ho2021classifier} & 4.703 (0.894) & 2.070 (0.022) & 0.923 (0.119) & 56.663 (39.914) \\
					\cdashline{2-8}
					\specialrule{0em}{1pt}{1pt}
					& \multirow{4}[0]{*}{Continuous} & \multirow{2}[0]{*}{GAN} & CcGAN (T-PAMI'23)~\cite{ding2023ccgan} & 1.334 (0.531) & 1.784 (0.065) & {1.234 (0.209)} & 14.807 (14.297) \\
					&    &   & Dual-NDA (AAAI'24)~\cite{ding2024turning} & {1.114 (0.503)} & \textbf{1.738 (0.055)} & \textbf{1.251 (0.172)} & {11.809 (11.694)} \\
					\cdashline{3-8}
					\specialrule{0em}{1pt}{1pt}
					&    & \multirow{2}[0]{*}{Diffusion}  & CcDPM (AAAI'24)~\cite{zhao2024ccdpm} & 0.939 (0.508)  & 1.761 (0.057)  & 1.150  (0.208)  &  10.999 (12.455) \\
					&    &   & \textbf{CCDM (ours)} & \textbf{0.742 (0.356)} & {1.778 (0.067)} & 1.088 (0.180) & \textbf{5.823 (5.213)} \\
					\midrule
					
					\multirow{8}[2]{*}{\begin{tabular}[c]{@{}c@{}} \textbf{Cell-200} \\ {(64$\times$64)}\end{tabular}} & \multirow{4}[0]{*}{\begin{tabular}[c]{@{}c@{}} Categorical\\ {(100 classes)}\end{tabular}} & \multirow{2}[0]{*}{GAN} & ReACGAN (Neurips'21)~\cite{kang2021rebooting} & 12.901 (4.099) & 1.652 (0.691) & NA    & 30.752 (30.710) \\
					&    &   & ADCGAN (ICML'22)~\cite{hou2022conditional} & 121.857 (11.744) & 4.411 (0.610) & NA     & 80.344 (54.579) \\
					\cdashline{3-8}
					\specialrule{0em}{1pt}{1pt}
					&    & \multirow{2}[0]{*}{Diffusion} & ADM-G (Neurips'21)~\cite{dhariwal2021diffusion} & 7.809 (4.701) & 3.280 (1.624) &  NA   & 31.365 (40.055) \\
					&    &   & CFG (Neurips'21)~\cite{ho2021classifier} & 8.236 (7.177) & 1.365 (0.399) & NA     & 21.176 (10.194) \\
					\cdashline{2-8}
					\specialrule{0em}{1pt}{1pt}
					& \multirow{4}[0]{*}{Continuous} &  \multirow{2}[0]{*}{GAN} & CcGAN (T-PAMI'23)~\cite{ding2023ccgan} & 6.848 (3.157) & 1.298 (0.576) & NA     & {5.210 (4.994)} \\
					&    &   & Dual-NDA (AAAI'24)~\cite{ding2024turning} & {6.439 (1.633)} & \textbf{1.095 (0.517)} & NA   & 6.420 (6.334) \\
					\cdashline{3-8}
					\specialrule{0em}{1pt}{1pt}
					&    & \multirow{2}[0]{*}{Diffusion}  & CcDPM (AAAI'24)~\cite{zhao2024ccdpm} & 6.394 (10.625) & 1.208 (0.505) & NA &  3.196 (3.513) \\
					&    &   & \textbf{CCDM (ours)} & \textbf{5.122 (1.433)} & {1.184 (0.578)} &  NA  & \textbf{2.941 (2.651)} \\
					\midrule
					
					\multirow{8}[2]{*}{\begin{tabular}[c]{@{}c@{}} \textbf{UTKFace} \\ {(128$\times$128)}\end{tabular}} & \multirow{4}[0]{*}{\begin{tabular}[c]{@{}c@{}} Categorical\\ {(60 classes)}\end{tabular}} & \multirow{2}[0]{*}{GAN} & ReACGAN (Neurips'21)~\cite{kang2021rebooting} & {0.445 (0.098)} & {1.426 (0.064)} & {1.152 (0.304)} & {6.005 (5.182)} \\
					&   &    & ADCGAN (ICML'22)~\cite{hou2022conditional} & {0.468 (0.143)} & {1.231 (0.048)} & \textbf{1.365 (0.035)} & {15.777 (11.572)} \\
					\cdashline{3-8}
					\specialrule{0em}{1pt}{1pt}
					&   &  \multirow{2}[0]{*}{Diffusion}  & ADM-G (Neurips'21)~\cite{dhariwal2021diffusion} & {0.997 (0.208)} & {3.705 (0.409)} & {0.831 (0.271)} & {11.618 (8.754)} \\
					&   &    & CFG (Neurips'21)~\cite{ho2021classifier} & {1.289 (0.261)} & {1.265 (0.116)} & {1.119 (0.228)} & {9.146 (7.591)} \\
					\cdashline{2-8}
					\specialrule{0em}{1pt}{1pt}
					
					& Text & Diffusion & SD v1.5 (CVPR'22)~\cite{rombach2022high} & 0.768 (0.212) & 2.412 (0.749) & 1.059 (0.353) & \textbf{5.195 (3.946)} \\
					\cdashline{2-8}
					\specialrule{0em}{1pt}{1pt}
					
					& \multirow{4}[0]{*}{Continuous} & \multirow{2}[0]{*}{GAN} & CcGAN (T-PAMI'23)~\cite{ding2023ccgan} & {0.367 (0.123)} & {1.113 (0.033)} & {1.199 (0.232)} & {7.747 (6.580)} \\
					&   &    & Dual-NDA (AAAI'24)~\cite{ding2024turning} & {0.361 (0.127)} & {1.081 (0.042)} & {1.257 (0.238)} & {6.310 (5.194)} \\
					\cdashline{3-8}
					\specialrule{0em}{1pt}{1pt}
					&    & \multirow{2}[0]{*}{Diffusion}  & CcDPM (AAAI'24)~\cite{zhao2024ccdpm} & 0.529 (0.118) & 1.114 (0.040) & 1.195 (0.197)  & 7.933 (6.716)  \\
					&   &    & \textbf{CCDM (ours)} & \textbf{0.319 (0.123)} & \textbf{1.077 (0.042)} & {1.178 (0.240)} & {6.359 (5.328)} \\
					\midrule
					
					\multirow{8}[2]{*}{\begin{tabular}[c]{@{}c@{}} \textbf{Steering Angle} \\ {(128$\times$128)}\end{tabular}} & \multirow{4}[0]{*}{\begin{tabular}[c]{@{}c@{}} Categorical\\ {(221 classes)}\end{tabular}} & \multirow{2}[0]{*}{GAN} & ReACGAN (Neurips'21)~\cite{kang2021rebooting} & 3.979 (0.919) & {4.060 (0.643)} & {0.250 (0.269)} & {36.631 (38.592)} \\
					&   &    & ADCGAN (ICML'22)~\cite{hou2022conditional} & {3.110 (0.799)} & {5.181 (0.010)} & {0.001 (0.001)} & {44.242 (29.223)} \\
					\cdashline{3-8}
					\specialrule{0em}{1pt}{1pt}
					&   &  \multirow{2}[1]{*}{Diffusion} & ADM-G (Neurips'21)~\cite{dhariwal2021diffusion} & {1.593 (0.449)} & {3.476 (0.153)} & {1.120 (0.121)} & {32.040 (27.836)} \\
					&   &    & CFG (Neurips'21)~\cite{ho2021classifier} & {5.425 (1.573)} & {2.742 (0.109)} & {0.762 (0.121)} & {50.015 (34.640)} \\
					\cdashline{2-8}
					\specialrule{0em}{1pt}{1pt}
					& \multirow{4}[0]{*}{Continuous} & \multirow{2}[0]{*}{GAN} & CcGAN (T-PAMI'23)~\cite{ding2023ccgan} & {1.689 (0.443)} & {2.411 (0.100)} & {1.088 (0.243)} & {18.438 (16.072)} \\
					&   &    & Dual-NDA (AAAI'24)~\cite{ding2024turning} & {1.390 (0.421)} & {2.135 (0.065)} & {1.133 (0.217)} & {14.099 (12.097)} \\
					\cdashline{3-8}
					\specialrule{0em}{1pt}{1pt}
					&    & \multirow{2}[0]{*}{Diffusion}  & CcDPM (AAAI'24)~\cite{zhao2024ccdpm} & 1.285 (0.520)  & 1.989 (0.055)  & \textbf{1.203 (0.191)}   &  18.325 (19.489)  \\
					&   &    & \textbf{CCDM (ours)} & \textbf{0.987 (0.419)} & \textbf{1.977 (0.062)} & 1.118 (0.225) & \textbf{11.829 (12.523)} \\
					\midrule
					
					
					\multirow{4}[2]{*}{\begin{tabular}[c]{@{}c@{}} \textbf{UTKFace} \\ {(192$\times$192)}\end{tabular}}& Text & Diffusion & SD v1.5 (CVPR'22)~\cite{rombach2022high} & 0.928 (0.284) & 2.941 (0.953) & 1.053 (0.321) & \textbf{5.065 (3.867)} \\
					\cdashline{2-8}
					\specialrule{0em}{1pt}{1pt}
					
					& \multirow{4}[0]{*}{Continuous} & \multirow{2}[0]{*}{GAN} &  CcGAN (T-PAMI'23)~\cite{ding2023ccgan} & {0.499 (0.186)} & {1.661 (0.047)} & \textbf{1.207 (0.260)} & {7.885 (6.272)} \\
					&   &   & Dual-NDA (AAAI'24)~\cite{ding2024turning} & {0.487 (0.179)} & {1.483 (0.045)} & 1.201 (0.258) & {6.730 (5.397)} \\
					\cdashline{3-8}
					\specialrule{0em}{1pt}{1pt}
					&    & \multirow{2}[0]{*}{Diffusion}  & CcDPM (AAAI'24)~\cite{zhao2024ccdpm} &  0.970 (0.192)  &  1.522 (0.066)  &  1.187 (0.152)  & 11.224 (8.827)  \\
					&   &    & \textbf{CCDM (ours)} &  \textbf{0.467 (0.178)}    &   \textbf{1.242 (0.062)}    &  1.148 (0.222)   &  7.336 (5.932) \\
					\bottomrule
				\end{tabular}%
				\begin{tablenotes}
					\scriptsize	
					\item \textit{In the RC-49, UTKFace, Steering Angle, and Cell-200 experiments, we generate 179,800, 60,000, 100,000, and 200,000 fake images, respectively, using each candidate method, with CCDM utilizing the DDIM sampler. Fake images with a resolution of $192\times 192$ are exclusively generated by the CCGM models. These generated images are assessed using four metrics: SFID, NIQE, Diversity, and Label Score. \textbf{We also report the standard deviations of the FID, NIQE, and Diversity scores for each method across predefined evaluation centers, as well as the standard deviation of the Label Score across all generated images in parentheses.} More details on these evaluation metrics can be found in Appendix~\ref{sec:supp_detailed_test_setups}. ``$\downarrow$" (``$\uparrow$") indicates lower (higher) values are preferred. The best results are marked in bold.}
				\end{tablenotes}
				\label{tab:main_results}%
			\end{threeparttable}
		\end{adjustbox}
	\end{table*}%
	
	\begin{table}[!htbp]
		\centering
		\caption{\revision{Comparison of Sampling Time and GPU Memory Usage Across CCGM Methods on RC-49. Evaluated on one NVIDIA RTX 4090D with Total Sample Size 179,800 and Fixed Batch Size 200.}}
		\begin{adjustbox}{width=0.8\linewidth}
			\begin{tabular}{cccc}
				\toprule
				\textbf{Method} & \textbf{SFID}  & \begin{tabular}[c]{@{}c@{}} {\textbf{Time}} \\ \textbf{(hours)} \end{tabular} & \begin{tabular}[c]{@{}c@{}} {\textbf{GPU Memory}} \\ \textbf{Usage (GiB)} \end{tabular}   \\
				\midrule
				CcGAN & 0.126 & 0.008 & 2.62 \\
				Dual-NDA & 0.148 & 0.008 & 2.62 \\
				CcDPM ($T^\prime=250$) & 0.970 & 17.05 & 5.07 \\
				\cdashline{1-4}
				\specialrule{0em}{1pt}{1pt}
				CCDM ($T^\prime=250$) & 0.049 & 17.05 & 5.53 \\
				DMD2-M & 0.096 & 0.009 & 2.44 \\
				\bottomrule
			\end{tabular}%
		\end{adjustbox}
		\label{tab:ab_sampling_cost}%
	\end{table}%

	
	\begin{table}[!htbp]
		\centering
		\caption{Comparison of Image Quality and Sampling Time between DMD2-M and Other Samplers in Two $64 \times 64$ Experiments. Sampling times were recorded on a platform equipped with an NVIDIA RTX 4090D. DDPM1K and DDPM250 denote the DDPM sampler with 1000 and 250 sampling steps, respectively, while DDIM250 refers to the DDIM sampler with 250 sampling steps.}
		
		\begin{adjustbox}{width=1\linewidth}
			\begin{tabular}{ccccccc}
				\toprule
				\textbf{Dataset} & \textbf{Sampler} & \textbf{SFID} & \textbf{NIQE}  & \textbf{Diversity} & \begin{tabular}[c]{@{}c@{}} \textbf{Label} \\ \textbf{Score} \end{tabular} & \begin{tabular}[c]{@{}c@{}} {\textbf{Time}} \\ \textbf{(hours)} \end{tabular} \\
				\midrule
				\multirow{4}[0]{*}{RC-49} & DDPM1K &   0.050 & 2.137 & 3.692  & 0.915 & 68.20 \\
				& DDPM250 & 0.749 & 1.849 & 2.308 & 2.679 & 17.05 \\
				& DDIM250 & 0.049 & 2.086 & 3.698 & 1.074 & 17.05 \\
				& \textbf{DMD2-M}  &  0.096 & 1.768 & 3.520 & 2.109 & 0.009 \\
				\cdashline{1-7}
				\specialrule{0em}{1pt}{1pt}
				
				\multirow{4}[0]{*}{\begin{tabular}[c]{@{}c@{}} Steering \\ Angle \end{tabular}} & DDPM1K &  0.732  &  1.776   &  1.051   &  5.457 &  33.044 \\
				& DDPM250 & 3.427  & 1.883 & 0.128 & 49.463    &  8.261 \\
				& DDIM250 & 0.742  & 1.778  & 1.088 & 5.823    &  8.261 \\
				& \textbf{DMD2-M} & 0.981 & 1.806 &  1.163  & 8.960 &  0.004 \\
				
				\bottomrule
			\end{tabular}%
		\end{adjustbox}
		\label{tab:ab_effect_samplers}%
	\end{table}%

	\section{Ablation Study}\label{sec:ablation}
	
\begin{figure}[!htbp] 
		\centering
		\begin{minipage}{0.7\linewidth}
			\centering
			\includegraphics[width=\linewidth]{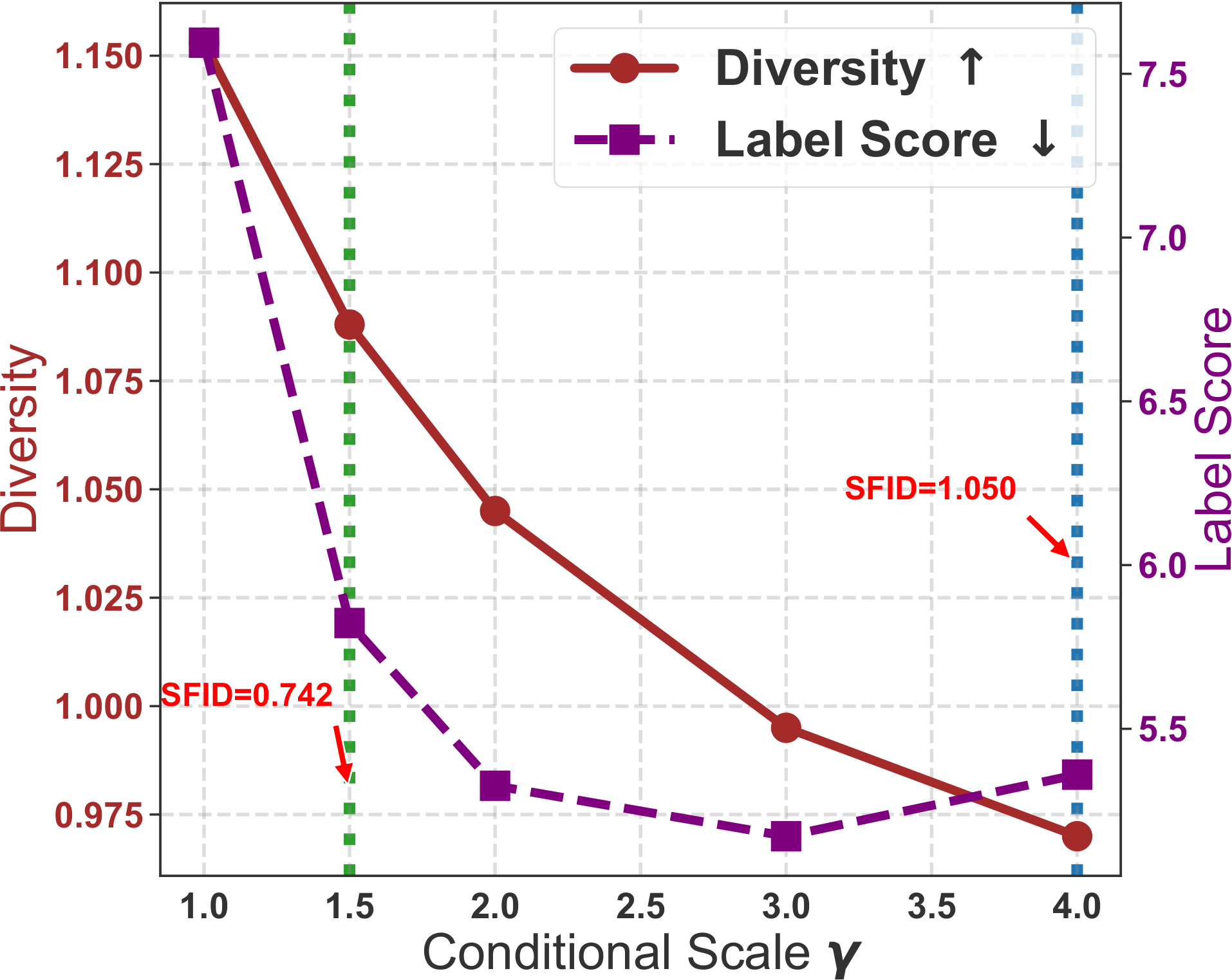}
		\end{minipage}
		\\
		\begin{minipage}{0.7\linewidth}
			\centering
			\includegraphics[width=\linewidth]{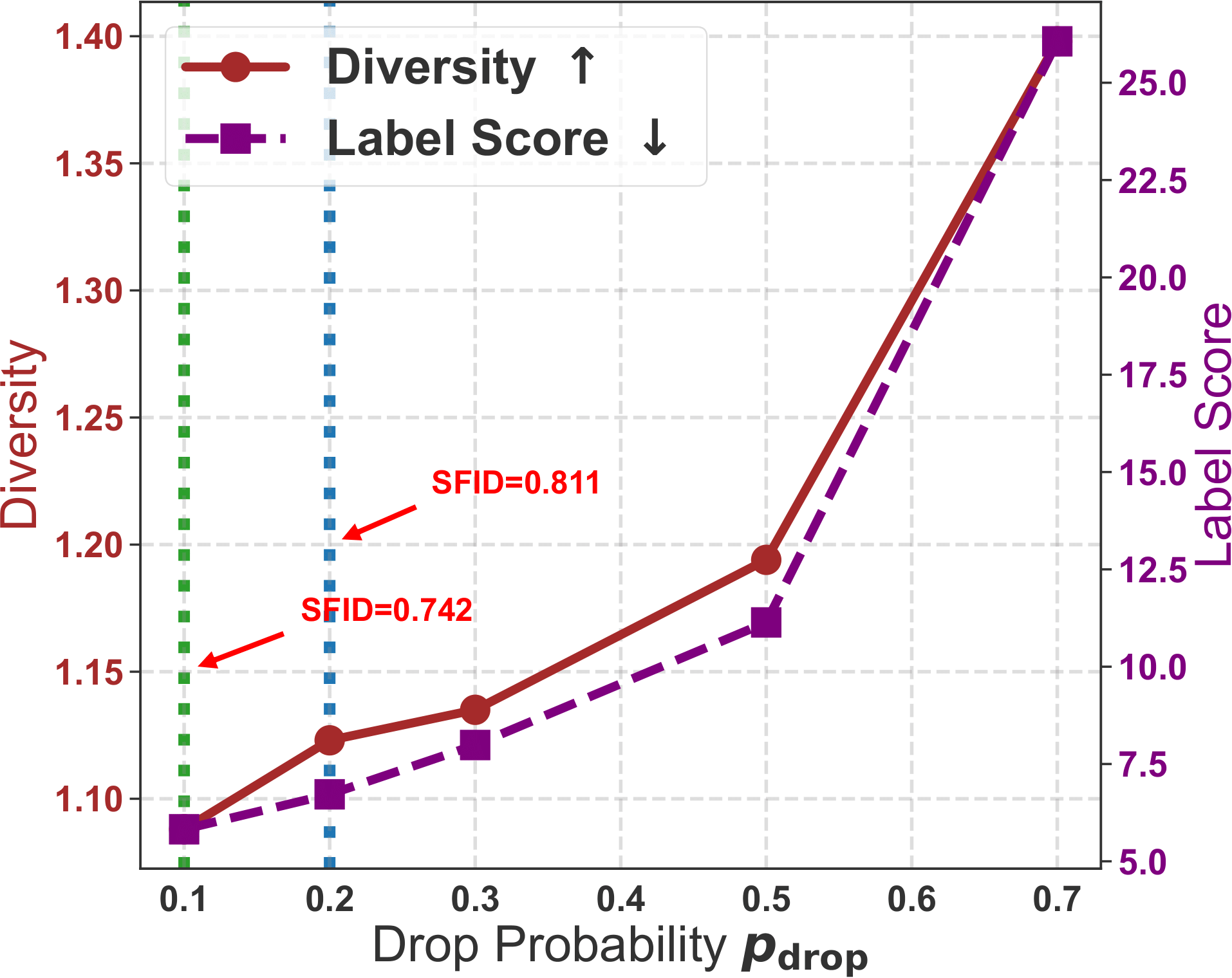}
		\end{minipage}
		\caption{\textit{Effects of $\gamma$ and $p_{\text{drop}}$ on diversity-label consistency trade-off for Steering Angle.} The green and blue vertical lines indicate the proposed values for CCGM and the commonly used values for discrete conditions, respectively.}
		\label{fig:ab_cond_drop_and_cond_scale}
	\end{figure}

	In addition to the experiments in Section~\ref{sec:experiment}, we also conduct eight ablation studies at a resolution of $64\times 64$ to evaluate CCDM's performance across different configurations, altering one setup or component at a time while keeping others constant. We examine the effects of the $y$-dependent covariance matrix (Table \ref{tab:ab_effect_Hy}), vicinity type (Table \ref{tab:ab_effect_vicinity}), training objective (Table \ref{tab:ab_effect_objective}), conditional scale $\gamma$ (Table \ref{tab:ab_effect_cond_scale}), condition drop probability $p_{\text{drop}}$ (Table \ref{tab:ab_effect_cond_drop_prob}), condition embedding method (Table \ref{tab:ab_effect_embedding}), sampling time steps $T^\prime$ (Table \ref{tab:ab_effect_sampling_timesteps}), and individual components of DMD2-M (Table \ref{tab:ablation_dmd2m}). These ablation studies collectively validate the effectiveness of our proposed configurations. Notably, Tables \ref{tab:ab_effect_vicinity} and \ref{tab:ab_effect_objective} show that the soft vicinity and $\bm{\epsilon}$-prediction loss in CcDPM are suboptimal for the CCGM task, with $\bm{x}^0$- and $\bm{v}$-prediction losses \cite{salimans2022progressive} performing similarly. Tables \ref{tab:ab_effect_cond_scale} and \ref{tab:ab_effect_cond_drop_prob} and Fig.~\ref{fig:ab_cond_drop_and_cond_scale} show that $\gamma$ and $p_{\text{drop}}$ balance the trade-off between diversity and label consistency, consistent with \cite{ho2021classifier}, though we recommend $\gamma \in [1.5, 2]$ and $p_{\text{drop}} = 0.1$ for optimal performance, in contrast to their suggestion of $\gamma > 3$ and $p_{\text{drop}} \geq 0.2$ for class-conditional scenarios. Additionally, experiments with two alternative methods for encoding regression labels—sinusoidal positional embedding \cite{ho2020denoising} and Gaussian Fourier projection \cite{song2021scorebased}—show poor label consistency, leading to suboptimal performance, as detailed in Table \ref{tab:ab_effect_embedding}. Finally, Table \ref{tab:ab_effect_sampling_timesteps} shows that reducing $T^\prime$ below 150 significantly degrades performance. Therefore, in $128\times 128$ and $192\times 192$ experiments, we set $T^\prime = 150$ to balance sampling time and image quality.

	\begin{table}[!htbp]
		\centering
		\caption{Performance Evaluation of CCDM Utilizing $y$-dependent vs. $y$-independent Covariance Matrices in $64\times 64$ Experiments.}
		\begin{adjustbox}{width=0.95\linewidth}
			\begin{tabular}{cccccc}
				\toprule
				\textbf{Dataset} & \begin{tabular}[c]{@{}c@{}} \textbf{Covariance} \\ \textbf{Matrix} \end{tabular} & \textbf{SFID} & \textbf{NIQE}  & \textbf{Diversity} & \begin{tabular}[c]{@{}c@{}} \textbf{Label} \\ \textbf{Score} \end{tabular}  \\
				\midrule
				\multirow{2}[0]{*}{RC-49} & $y$-independent & {0.050} & {2.104} & {3.697} & {1.081} \\
				& $y$-dependent & \textbf{0.049} & \textbf{2.086} & \textbf{3.698} & \textbf{1.074} \\
				\cdashline{1-6}
				\specialrule{0em}{1pt}{1pt}
				
				\multirow{2}[0]{*}{UTKFace} & $y$-independent  & 0.370 & 1.549 & 1.181 & 6.176 \\
				& $y$-dependent & \textbf{0.363} & \textbf{1.542} & \textbf{1.184} & \textbf{6.164} \\
				\cdashline{1-6}
				\specialrule{0em}{1pt}{1pt}
				
				\multirow{2}[0]{*}{Cell-200} & $y$-independent  & 5.183 & 1.229 & NA    & 3.036 \\
				& $y$-dependent & \textbf{5.122} & \textbf{1.184} & NA    & \textbf{2.941} \\
				\cdashline{1-6}
				\specialrule{0em}{1pt}{1pt}
				
				\multirow{2}[0]{*}{\begin{tabular}[c]{@{}c@{}} Steering \\ Angle \end{tabular}} & $y$-independent  & 0.751 & 1.780 & \textbf{1.118} & 5.973 \\
				& $y$-dependent & \textbf{0.742} & \textbf{1.778} & 1.088 & \textbf{5.823} \\
				\bottomrule
			\end{tabular}%
		\end{adjustbox}
		\label{tab:ab_effect_Hy}%
	\end{table}%
	
	\begin{table}[!htbp]
		\centering
		\caption{Performance Evaluation of CCDM with Varied Vicinity Types in $64\times 64$ Experiments.}
		\begin{adjustbox}{width=0.85\linewidth}
			\begin{tabular}{cccccc}
				\toprule
				\textbf{Dataset} & \begin{tabular}[c]{@{}c@{}} \textbf{Vicinity} \\ \textbf{Type} \end{tabular} & \textbf{SFID} & \textbf{NIQE}   & \textbf{Diversity} & \begin{tabular}[c]{@{}c@{}} \textbf{Label} \\ \textbf{Score} \end{tabular} \\
				\midrule
				\multirow{3}[0]{*}{RC-49 } & None & 0.051 & 2.056 & 3.692 & 1.104 \\
				& Soft & 0.053 & \textbf{1.986} & 3.679 & 1.337 \\
				& Hard & \textbf{0.049} & 2.086 & \textbf{3.698} & \textbf{1.074} \\
				\cdashline{1-6}
				\specialrule{0em}{1pt}{1pt}
				
				\multirow{3}[0]{*}{UTKFace} & None & 0.375 & 1.548 & \textbf{1.184} & 6.312 \\
				& Soft & 0.375  & 1.545 & 1.179 & 6.380 \\
				& Hard & \textbf{0.363} & \textbf{1.542} & \textbf{1.184} & \textbf{6.164} \\
				\cdashline{1-6}
				\specialrule{0em}{1pt}{1pt}
				
				\multirow{3}[0]{*}{Cell-200} & None & 5.276 & 1.204 & NA    & 3.013 \\
				& Soft & 5.411 & 1.226 & NA    & 3.152 \\
				& Hard & \textbf{5.122} & \textbf{1.184} & NA    & \textbf{2.941} \\
				\cdashline{1-6}
				\specialrule{0em}{1pt}{1pt}
				
				\multirow{3}[0]{*}{\begin{tabular}[c]{@{}c@{}} Steering \\ Angle \end{tabular}} & None & 0.922 & \textbf{1.758} & \textbf{1.127} & 7.376 \\
				& Soft & 0.848 & 1.773 & 1.097 & 6.652 \\
				& Hard & \textbf{0.742} & 1.778  & 1.088 & \textbf{5.823} \\
				\bottomrule
			\end{tabular}%
		\end{adjustbox}
		\label{tab:ab_effect_vicinity}%
	\end{table}%
	
	\begin{table}[!htbp]
		\centering
		\caption{Performance Evaluation of CCDM with Varied Training Objectives in $64\times 64$ Experiments.}
		\begin{adjustbox}{width=0.9\linewidth}
			\begin{tabular}{cccccc}
				\toprule
				\textbf{Dataset} & \begin{tabular}[c]{@{}c@{}} \textbf{Training} \\ \textbf{Objective} \end{tabular} & \textbf{SFID}  & \textbf{NIQE}   & \textbf{Diversity} & \begin{tabular}[c]{@{}c@{}} \textbf{Label} \\ \textbf{Score} \end{tabular} \\
				\midrule
				\multirow{3}[0]{*}{RC-49 } & $\bm{\epsilon}$-pred. & 0.560 & 2.164 & 3.607 & 16.675 \\
				& $\bm{v}$-pred. & 0.048 & 1.856 & 3.695 & 0.928 \\
				& $\bm{x}^0$-pred. & 0.049 & 2.086 & 3.698 & 1.074 \\
				\cdashline{1-6}
				\specialrule{0em}{1pt}{1pt}
				
				\multirow{3}[0]{*}{UTKFace} & $\bm{\epsilon}$-pred. & 0.444 & 1.539 & 1.199 & 6.586 \\
				& $\bm{v}$-pred. & 0.362  & 1.543 & 1.178 & 6.074 \\
				& $\bm{x}^0$-pred. & 0.363 & 1.542 & 1.184 & 6.164 \\
				\cdashline{1-6}
				\specialrule{0em}{1pt}{1pt}
				
				\multirow{3}[0]{*}{Cell-200} & $\bm{\epsilon}$-pred. & 5.866 & 1.153 & NA    & 2.688 \\
				& $\bm{v}$-pred. & 5.202  & 1.186 & NA    & 2.631 \\
				& $\bm{x}^0$-pred. & 5.122 & 1.184 & NA    & 2.941 \\
				\cdashline{1-6}
				\specialrule{0em}{1pt}{1pt}
				
				\multirow{3}[0]{*}{\begin{tabular}[c]{@{}c@{}} Steering \\ Angle \end{tabular}} & $\bm{\epsilon}$-pred. & 0.867 & 1.766 & 1.135 & 10.029 \\
				& $\bm{v}$-pred. & 0.775  & 1.781  & 1.101 & 6.025 \\
				& $\bm{x}^0$-pred. & 0.742 & 1.778  & 1.088 & 5.823 \\
				\bottomrule
			\end{tabular}%
		\end{adjustbox}
		\label{tab:ab_effect_objective}%
	\end{table}%

	\begin{table}[htbp]
		\centering
		\caption{Performance Evaluation of CCDM with Varied Conditional Scale $\gamma$ in Two $64\times 64$ Experiments.}
		\begin{adjustbox}{width=0.77\linewidth}
			\begin{tabular}{cccccc}
				\toprule
				\textbf{Dataset} & $\gamma$  & \textbf{SFID} & \textbf{NIQE} & \textbf{Diversity} & \begin{tabular}[c]{@{}c@{}} \textbf{Label} \\ \textbf{Score} \end{tabular} \\
				\midrule
				\multirow{5}[1]{*}{RC-49} & 4.0   & 0.112 & 1.969 & 3.561 & 0.801 \\
				& 3.0   & 0.088 & 2.001 & 3.610  & 0.839 \\
				& 2.0   & 0.061 & 2.050 & 3.669 & 0.950 \\
				& 1.5   & 0.049 & 2.086 & 3.698 & 1.074 \\
				& 1.0   & 0.046 & 2.108 & 3.712 & 1.498 \\
				\cdashline{1-6}
				\specialrule{0em}{1pt}{1pt}
				
				\multirow{5}[0]{*}{\begin{tabular}[c]{@{}c@{}} Steering \\ Angle \end{tabular}} & 4.0   & 1.050 & 1.762 & 0.970 & 5.360 \\
				& 3.0   & 0.922 & 1.772 & 0.995 & 5.172 \\
				& 2.0   & 0.792 & 1.776 & 1.045 & 5.326  \\
				& 1.5   & 0.742 & 1.778 & 1.088 & 5.823  \\
				& 1.0   & 0.755 & 1.776 & 1.153 & 7.595  \\
				\bottomrule
			\end{tabular}%
		\end{adjustbox}
		\label{tab:ab_effect_cond_scale}%
	\end{table}%
	
	\begin{table}[!htbp]
		\centering
		\caption{Performance Evaluation of CCDM with Varied Condition Drop Probability $p_{\text{drop}}$ in Two $64\times 64$ Experiments.}
		\begin{adjustbox}{width=0.8\linewidth}
			\begin{tabular}{cccccc}
				\toprule
				\textbf{Dataset} & $p_{\text{drop}}$ & \textbf{SFID} & \textbf{NIQE} & \textbf{Diversity} & \begin{tabular}[c]{@{}c@{}} \textbf{Label} \\ \textbf{Score} \end{tabular} \\
				\midrule
				\multirow{4}[0]{*}{RC-49} & 0.1 & 0.049 & 2.086 & 3.698 & 1.074 \\
				& 0.2   & 0.050 & 2.017 & 3.698 & 1.398  \\
				& 0.3   & 0.052 & 2.042 & 3.694  & 1.650 \\
				& 0.5   & 0.059 & 2.008 & 3.695  & 2.860  \\
				& 0.7   & 0.104 & 2.148 & 3.684 & 5.118 \\
				\cdashline{1-6}
				\specialrule{0em}{1pt}{1pt}
				
				\multirow{5}[0]{*}{\begin{tabular}[c]{@{}c@{}} Steering \\ Angle \end{tabular}} & 0.1   & 0.742 & 1.778 & 1.088 & 5.823 \\
				& 0.2   & 0.811 & 1.789 & 1.123 & 6.721 \\
				& 0.3   & 0.894 & 1.786 & 1.135 & 7.987  \\
				& 0.5   & 1.089 & 1.783 & 1.194 & 11.137 \\
				& 0.7   & 1.506 & 1.781 & 1.398 & 26.064 \\
				\bottomrule
			\end{tabular}%
		\end{adjustbox}
		\label{tab:ab_effect_cond_drop_prob}%
	\end{table}%

	
	\begin{table}[!htbp]
		\centering
		\caption{Performance Evaluation of CCDM with Varied Condition Embedding Methods in Two $64\times 64$ Experiments.}
		\begin{adjustbox}{width=0.9\linewidth}
			\begin{tabular}{cccccc}
				\toprule
				\textbf{Dataset} & \begin{tabular}[c]{@{}c@{}} \textbf{Embedding} \\ \textbf{Method} \end{tabular} & \textbf{SFID} & \textbf{NIQE}  & \textbf{Diversity} & \begin{tabular}[c]{@{}c@{}} \textbf{Label} \\ \textbf{Score} \end{tabular}  \\
				\midrule
				\multirow{3}[3]{*}{RC-49} & Sinusoidal & 0.469 & 2.178 & 3.643 & 14.815 \\
				& \begin{tabular}[c]{@{}c@{}} {Gaussian} \\ {Fourier} \end{tabular} & 0.859 & 2.186  & \textbf{3.724} & 29.692  \\
				& CNN (ours)   & \textbf{0.049} & \textbf{2.086} & 3.698 & \textbf{1.074} \\
				\cdashline{1-6}
				\specialrule{0em}{1pt}{1pt}
				
				\multirow{3}[3]{*}{\begin{tabular}[c]{@{}c@{}} Steering \\ Angle \end{tabular}} & Sinusoidal & 1.129  & 1.758  & 1.222 & 11.118 \\
				& \begin{tabular}[c]{@{}c@{}} {Gaussian} \\ {Fourier} \end{tabular} & 1.924  & \textbf{1.763} & \textbf{1.489} & 50.922 \\
				& CNN (ours) & \textbf{0.742} & 1.778 & 1.088 & \textbf{5.823} \\
				\bottomrule
			\end{tabular}%
		\end{adjustbox}
		\label{tab:ab_effect_embedding}%
	\end{table}%
	
	
	\begin{table}[!htbp]
		\centering
		\caption{Evaluation of CCDM Performance Using DDIM Sampler with Varying Time Steps $T^\prime$ in Two $64\times 64$ Experiments.}
		\begin{adjustbox}{width=0.85\linewidth}
			\begin{tabular}{cccccc}
				\toprule
				\textbf{Dataset} & \begin{tabular}[c]{@{}c@{}} \textbf{Time} \\ \textbf{Steps} $T^\prime$ \end{tabular} & \textbf{SFID}  & \textbf{NIQE}  & \textbf{Diversity} & \begin{tabular}[c]{@{}c@{}} \textbf{Label} \\ \textbf{Score} \end{tabular} \\
				\midrule
				\multirow{5}[0]{*}{RC-49} & 250   & 0.049 & 2.086 & 3.698 & 1.074 \\
				& 200   & 0.049 & 2.083 & 3.699 & 1.078 \\
				& 150 & 0.048   &  2.078  & 3.699  & 1.086   \\
				& 100   & 0.050  & 2.075  & 3.698  & 1.091  \\
				& 50    & 0.052 & 2.048 & 3.694 & 1.115 \\
				\cdashline{1-6}
				\specialrule{0em}{1pt}{1pt}
				
				\multirow{5}[0]{*}{\begin{tabular}[c]{@{}c@{}} Steering \\ Angle \end{tabular}}  & 250   & 0.742 & 1.778 & 1.088 & 5.823 \\
				& 200   & 0.747 & 1.778 & 1.091 & 5.844 \\
				& {150}  & {0.745} & {1.778} & {1.090} & {5.859} \\
				& 100   & 0.752 & 1.784 & 1.089 & 5.904 \\
				& 50    & 0.771 & 1.799 & 1.094 & 6.095 \\
				\bottomrule
			\end{tabular}%
		\end{adjustbox}
		\label{tab:ab_effect_sampling_timesteps}%
	\end{table}%

	\begin{table*}[!htbp]
		\centering
		\begin{adjustbox}{width=0.8\textwidth}
			\begin{threeparttable}
				\caption{Ablation Study on Individual Components of DMD2-M}
				\begin{tabular}{c|cccc|cccc}
					\toprule
					\multirow{2}[0]{*}{\textbf{Dataset}}  & \multicolumn{4}{c|}{\textbf{Configuration}} & \multirow{2}[0]{*}{\textbf{SFID}} & \multirow{2}[0]{*}{\textbf{NIQE}} & \multirow{2}[0]{*}{\textbf{Diversity}} & \multirow{2}[0]{*}{\begin{tabular}[c]{@{}c@{}} \textbf{Label} \\ \textbf{Score} \end{tabular}} \\
					\textbf{} & {SNGAN} & {Hinge Loss} & {DiffAugment} & {Hard Vicinal Weight} & \textbf{} & \textbf{} & \textbf{} & \textbf{} \\
					\midrule
					\multirow{5}[0]{*}{\begin{tabular}[c]{@{}c@{}} RC-49 \\ ($64\times 64$) \end{tabular}} &    &     &     &     & 2.312 &  1.617   & 0.000 &  39.695 \\
					& \checkmark  &      &    &   & 0.793 &  1.805   & 2.302 &  1.636 \\
					& \checkmark   &  \checkmark &  &  & 0.189 &  1.732   & 3.354 &  1.818 \\
					& \checkmark   & \checkmark   & \checkmark &   & 0.096 & 1.768 & 3.520 & 2.109  \\
					& \checkmark   & \checkmark   & \checkmark & \checkmark & 0.099 &  1.772   & 3.522 &  2.185  \\
					\midrule
					
					\multirow{5}[0]{*}{\begin{tabular}[c]{@{}c@{}} Steering Angle \\ ($64\times 64$) \end{tabular}} &    &     &     &     & 5.087 &  1.643 & 0.875  &  53.897 \\
					&  \checkmark  &   &     &     & 2.953 & 1.781 & 0.810 & 9.871 \\
					& \checkmark  &  \checkmark  &     &     & 4.094 & 2.135 & 0.372 & 11.735 \\
					& \checkmark &  \checkmark  & \checkmark   &     & 1.877 & 2.032 & 1.157 & 11.372 \\
					& \checkmark &  \checkmark   & \checkmark   & \checkmark   & 0.981 & 1.806 & 1.163 & 8.960 \\
					\bottomrule
				\end{tabular}%
				\begin{tablenotes}
					\scriptsize	
					\item \textit{The first row of each dataset represents the performance of the vanilla DMD2~\cite{yin2024improved}, while the following rows show the performance of DMD2-M. If the hinge loss option is not selected, the vanilla GAN loss is used. We recommend applying the hard vicinal weight in the Steering Angle experiment, where the training data is highly imbalanced, as DMD2-M collapses during training without it. Therefore, this table presents the evaluation performance of the last checkpoint before collapse under these settings. For RC-49, the hard vicinal weight does not significantly improve performance, which is why we set $\kappa=0$ for DMD2-M in Table \ref{tab:ab_effect_samplers}.}
				\end{tablenotes}
				\label{tab:ablation_dmd2m}%
			\end{threeparttable}
		\end{adjustbox}
	\end{table*}%

	\section{Conclusion}\label{sec:discussion}
	
	In this paper, we introduce CCDM, a novel CCGM model designed to generate high-quality images conditioned on regression labels. This is the first work that proposes conditional diffusion models for the CCGM task. Our approach meticulously tailors both the forward and reverse diffusion processes to incorporate regression labels explicitly. Additionally, we propose an effective covariance embedding network for encoding regression labels and introduce a specialized hard vicinal loss to train the denoising U-Net. To enhance sampling efficiency, we then develop two efficient conditional sampling procedures tailored for CCDM. Extensive experiments on four image datasets demonstrate that CCDM outperforms state-of-the-art conditional generative models in overall performance. \revision{As future work, we aim to incorporate more advanced diffusion frameworks to further enhance CCDM’s generalization and applicability.}

	\bibliographystyle{IEEEtran}
	\bibliography{./bibliography}

	\newpage

	\begin{center}
		\textbf{APPENDIX}
	\end{center}
	
	\renewcommand{\thesection}{S.\Roman{section}} 
	\renewcommand{\thesubsection}{\thesection.\Alph{subsection}}
	\renewcommand\thefigure{S.\arabic{figure}}
	\renewcommand\thetable{S.\arabic{table}}
	\renewcommand{\theequation}{S.\arabic{equation}}
	\renewcommand{\thetheorem}{S.\arabic{theorem}} 
	\renewcommand{\theremark}{S.\arabic{remark}}

	\renewcommand{\thealgocf}{S.\arabic{algocf}} 
	
	\setcounter{section}{0} 
	\setcounter{figure}{0} 
	\setcounter{table}{0} 
	\setcounter{equation}{0} 
	\setcounter{theorem}{0} 
	\setcounter{remark}{0} 
	
	\section{GitHub repository}\label{supp:codes}
	The codes and detailed setups can be found at
	\begin{center}
		\url{https://github.com/UBCDingXin/CCDM}
	\end{center}

	\section{The Network Design}\label{sec:supp_networks}
	
	\subsection{Architecture of Denoising U-Net}\label{sec:supp_unet_arch}
	
\begin{figure}[!htbp] 
		\centering
		\includegraphics[width=0.67\linewidth]{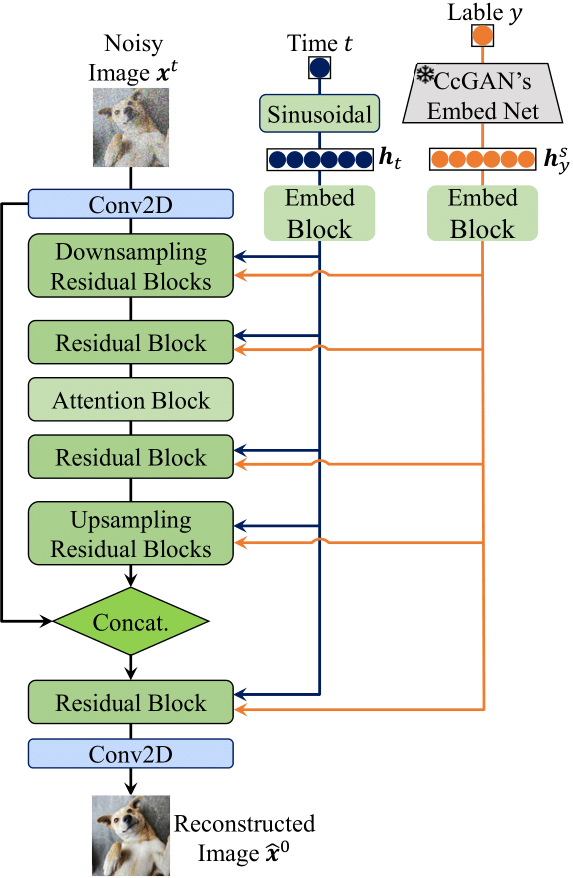}
		\caption{The network architecture of the denoising U-Net.}
		\label{fig:supp_unet_architecture}
	\end{figure}

	The denoising U-Net $\hat{\bm{x}}_{\bm{\theta}}^0(\bm{x}^t,t,y)$ is implemented based on Phil Wang's (lucidrain) GitHub repository\footnote{\url{https://github.com/lucidrains/denoising-diffusion-pytorch/blob/main/denoising_diffusion_pytorch/classifier_free_guidance.py}}. Its architecture is shown in Fig.~\ref{fig:supp_unet_architecture}. We adapt the original embedding block, initially designed for class labels, to incorporate the short embedding $\bm{h}_y^s$ from CcGAN's embedding network. The modified block consists of multiple fully-connected layers, each with 1D batch normalization and ReLU activation.
	\vspace{-0.5em}
	\vspace{-1em}
	
	\subsection{Covariance Embedding Network}\label{sec:supp_cov_label_embedding}
	
	Unlike DDPM's forward and reverse diffusion processes, the conditional diffusion processes in Section~\ref{sec:conditional_diffusion_process} explicitly incorporate the condition $y$. The covariance matrix of the transition $q(\bm{x}^t|\bm{x}^{t-1},y)$ now depends on $y$, with its diagonal corresponding to a high-dimensional embedding of the condition, denoted by $\bm{h}_y^l$. Notably, $\bm{h}_y^l$ must match the dimensionality of the flattened image $\bm{x}^0$, often making it much larger than $\bm{h}_y^s$ (typically 128), which is input to the denoising U-Net. To achieve this, we design a \textbf{covariance embedding network} to map the scalar $y$ to a high-dimensional vector $\bm{h}_y^l$. 
	
\begin{figure}[!htbp]
		\centering
		\includegraphics[width=0.75\linewidth]{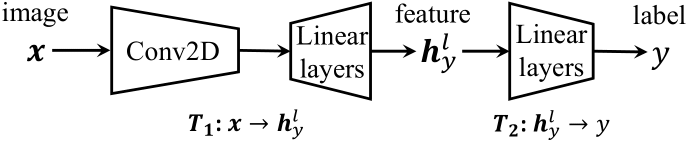}
		\caption{\textbf{The pre-trained CNN $T_1+T_2$ for the covariance label embedding.} The first subnetwork, $T_1$, includes convolutional layers (Conv2D) followed by linear layers, while the second subnetwork, $T_2$, consists solely of linear layers.}
		\label{fig:supp_pre_trained_CNN_for_label_embedding}
	\end{figure}

\begin{figure}[!htbp]
		\centering
		\includegraphics[width=0.65\linewidth]{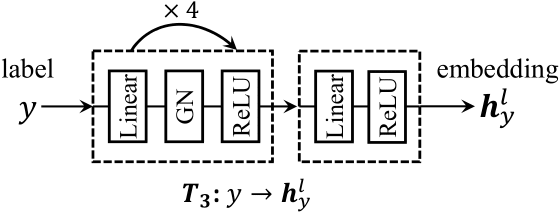}
		\caption{\textbf{The architecture of the covariance embedding network.} The covariance embedding network employs a 5-layer MLP, where the first four linear layers are each followed by group normalization (GN) and a ReLU activation.}
		\label{fig:supp_cov_label_embedding_network}
	\end{figure}

	Inspired by ILI’s label embedding network~\cite{ding2023ccgan}, we first pre-train a ResNet34, following the architecture in Fig.~\ref{fig:supp_pre_trained_CNN_for_label_embedding}, to predict regression labels from images. Unlike \cite{ding2023ccgan}, we train this model for only 10 epochs instead of 200, as its prediction performance does not impact label embedding. Within this ResNet, a linear layer expands hidden maps to ensure $\bm{h}_y^l$ matches $\bm{x}^0$'s dimension. As shown in Fig.~\ref{fig:supp_cov_label_embedding_network}, the covariance embedding network uses a 5-layer MLP (MLP-5) to map $y$ to $\bm{h}_y^l$. For $\tilde{N}$ distinct training labels $y_{[1]}, y_{[2]}, \dots, y_{[\tilde{N}]}$, the network $T_3$ is trained (with $T_1$ and $T_2$ fixed) using:
	\[
	\label{eq:supp_cov_embedding_loss}
	\min_{T_3}\frac{1}{\tilde{N}}\sum_{i=1}^{\tilde{N}}\mathbb{E}_{\zeta\sim\mathcal{N}(0,0.04)}\left[ (T_2(T_3(y_{[i]}+\zeta)) - (y_{[i]}+\zeta))^2 \right].
	\]
	\vspace{-1em}
	\vspace{-1em}
	
	\vspace{-1em}

	\section{Derivation of Reverse Diffusion Process and Hard Vicinal Image Denoising Loss}\label{sec:supp_detailed_derivations}

	\subsection{Derivation of Reverse Process}\label{sec:supp_derivation_reverse_process}
	
	The forward diffusion process has the subsequent transition:
	\[
	\label{eq:supp_forward_transition}
	q(\bm{x}^t|\bm{x}^{t-1},y)=\distNorm(\bm{x}^t; \sqrt{\alpha_t}\bm{x}^{t-1}, (1-\alpha_t) \bm{H}_y),
	\]
	where $\bm{H}_y=\text{diag}(\exp(-\bm{h}^l_y))$ is a $y$-dependent diagonal covariance matrix. Eq.~\eqref{eq:supp_forward_transition} differs from the transition in DDPM in that the identity covariance matrix $\bm{I}$ is substituted with $\bm{H}_y$. Due to the Markov property, we have:
	\[
	\label{eq:supp_forward_transition_gt}
	q(\bm{x}^t|\bm{x}^{t-1}, \bm{x}^0,y)=q(\bm{x}^t|\bm{x}^{t-1},y).
	\]
	
	Additionally, we know:
	\[
	\label{eq:supp_q_xt_x0}
	q(\bm{x}^t|\bm{x}^0,y)=\distNorm(\bm{x}^t; \sqrt{\bar{\alpha}_t}\bm{x}^0, (1-\bar{\alpha}_t) \bm{H}_y),
	\]
	and
	\[
	\label{eq:supp_q_xt_1_x0}
	q(\bm{x}^{t-1}|\bm{x}^0,y)=\distNorm(\bm{x}^{t-1}; \sqrt{\bar{\alpha}_{t-1}}\bm{x}^0, (1-\bar{\alpha}_{t-1}) \bm{H}_y).
	\]
	
	Using these, we can derive the ground truth denoising transition $q(\bm{x}^{t-1}|\bm{x}^t,\bm{x}^0,y)$ as follows:
	\[
	& q(\bm{x}^{t-1}|\bm{x}^t,\bm{x}^0,y) \\
	& = \frac{q(\bm{x}^t|\bm{x}^{t-1},\bm{x}^0,y)q(\bm{x}^{t-1}|\bm{x}^0,y)}{q(\bm{x}^t|\bm{x}^0,y)} \\
	& = \frac{\distNorm(\sqrt{\alpha_t}\bm{x}^{t-1}, (1-\alpha_t) \bm{H}_y) \distNorm(\sqrt{\bar{\alpha}_{t-1}}\bm{x}^0, (1-\bar{\alpha}_{t-1}) \bm{H}_y) }{ \distNorm(\sqrt{\bar{\alpha}_t}\bm{x}^0, (1-\bar{\alpha}_t) \bm{H}_y) } \\
	& = C_1 \exp\left\{ -\frac{1}{2} \left[  \frac{(\bm{x}^t-\sqrt{\alpha_t}\bm{x}^{t-1})^\intercal \bm{H}_y^{-1} (\bm{x}^t-\sqrt{\alpha_t}\bm{x}^{t-1}) }{1-\alpha_t}  \right. \right. \\
	& \qquad + \frac{(\bm{x}^{t-1}-\sqrt{\bar{\alpha}_{t-1}}\bm{x}^0)^\intercal \bm{H}_y^{-1} (\bm{x}^{t-1}-\sqrt{\bar{\alpha}_{t-1}}\bm{x}^0)}{1-\bar{\alpha}_{t-1}} \\
	& \left.\left. \qquad  + \frac{(\bm{x}^t-\sqrt{\bar{\alpha}_t}\bm{x}^0)^\intercal \bm{H}_y^{-1} (\bm{x}^t-\sqrt{\bar{\alpha}_t}\bm{x}^0)}{1-\bar{\alpha}_t} \right] \right\} \\
	& = C_1 \exp\left\{ -\frac{1}{2}\left[ \frac{-2\sqrt{\alpha_t}{\bm{x}^{t}}^\intercal\bm{H}_y^{-1}\bm{x}^{t-1}+\alpha_t{\bm{x}^{{t-1}}}^\intercal\bm{H}_y^{-1}\bm{x}^{t-1}   }{1-\alpha_t}  \right. \right. \\
	& \qquad \left.\left. + \frac{ {\bm{x}^{{t-1}}}^\intercal \bm{H}_y^{-1} \bm{x}^{t-1} - 2\sqrt{\bar{\alpha}_{t-1}}{\bm{x}^{{t-1}}}^\intercal  \bm{H}_y^{-1} \bm{x}^0    }{1-\bar{\alpha}_{t-1}}  + C_2 \right]\right\} \\
	& = C_1\exp\left\{\frac{1}{2}\left[ \frac{1-\bar{\alpha}_t}{(1-\alpha_t)(1-\bar{\alpha}_{t-1})} {\bm{x}^{{t-1}}}^\intercal \bm{H}_y^{-1} \bm{x}^{{t-1}}  \right.\right. \\ 
	& \left.\left. \qquad -2\left( \frac{\sqrt{\alpha_t}\bm{x}^t}{1-\alpha_t} + \frac{\sqrt{\bar{\alpha}_{t-1}}\bm{x}^0}{1-\bar{\alpha}_{t-1}} \right)^\intercal \bm{H}_y^{-1} \bm{x}^{{t-1}} +C_2  \right] \right\}\\
	&=C_1\exp\left\{ -\frac{1}{2\sigma^2_{q}(t)} \left( \bm{x}^{t-1}-\mu_q(\bm{x}^t,\bm{x}^0) \right)^\intercal \bm{H}_y^{-1}  \right. \\
	&\qquad\qquad \left.\cdot \left( \bm{x}^{t-1}-\mu_q(\bm{x}^t,\bm{x}^0,y) \right) \right\} \\
	&=\frac{1}{(2\pi)^\frac{d}{2}\lvert \sigma_q^2(t) \bm{H}_y \rvert^\frac{1}{2}} \exp\left\{ -\frac{1}{2} \left( \bm{x}^{t-1} - \mu_q(\bm{x}^t,\bm{x}^0,y) \right)^\intercal    \right. \\
	&\qquad\qquad \left. \cdot \left(\sigma_q^2(t)\bm{H}_y \right)^{-1}  \left( \bm{x}^{t-1} - \mu_q(\bm{x}^t,\bm{x}^0,y) \right) \right\} \\
	&=\distNorm(\bm{x}^{t-1}; \mu_q(\bm{x}^t,\bm{x}^0,y), \sigma_q^2(t)\bm{H}_y),
	\]
	where $C_1$ and $C_2$ absorb some constants that have no impact on the derivations, and $\mu_q(\bm{x}^t,\bm{x}^0,y)$ is defined as
	\[
	\bm{\mu}_q(\bm{x}^t,\bm{x}^0,y)= \frac{\sqrt{\alpha_t}(1-\bar{\alpha}_{t-1})\bm{x}^t+\sqrt{\bar{\alpha}_{t-1}}(1-\alpha_t)\bm{x}^0}{1-\bar{\alpha}_t}.
	\]

	\subsection{Derivation of Hard Vicinal Image Denoising Loss}\label{sec:supp_detailed_derivation_loss}
	
	CCDMs aim to estimate the conditional distribution $p(\bm{x}^0|y)$. Using $N$ image-label pairs $(\bm{x}^0_1,y_1),...,(\bm{x}^0_N,y_N)$ to learn the unknown parameters $\bm{\theta}$, the \textit{Negative Log-Likelihood} (NLL) of CCDM and its upper bound are typically defined as:
	\[
	&-\log\ex_{y\sim p(y)}\left[ \prod_{k=1}^{N^y} f_{\bm{\theta}}(\bm{x}_{k,y}^0|y) \right]\\
	\label{eq:supp_cond_loglik}
	\leq & -\ex_{y\sim p(y)} \left[ \log \prod_{k=1}^{N^y} f_{\bm{\theta}}(\bm{x}_{k,y}^0|y) \right] \quad\text{(By Jensen's inequality)} \\
	= & \ex_{y\sim p(y)} \left[ \sum_{i=1}^{N} \ind_{\{y_i=y\}} \left( -\log f_{\bm{\theta}}(\bm{x}_{i}^0|y) \right) \right] \label{eq:supp_loglik_lb}
	\]
	where $f_{\bm{\theta}}(\bm{x}^0|y)$ is the likelihood function for observed images with label $y$, $\bm{x}_{k,y}^0$ is the $k$-th image with label $y$, $N^y$ is the number of images with label $y$, $p(y)$ is the label distribution, and $\ind$ is an indicator function. However, Eq.~\eqref{eq:supp_loglik_lb} implies that estimating $p(\bm{x}^0|y)$ relies solely on images with label $y$, potentially suffering from the data insufficiency issue. 
	
	To address data sparsity, we integrate the vicinal loss from CcGANs into CDM training. By using training images with labels in a hard vicinity of $y$ to estimate $p(\bm{x}^0|y)$, we modify Eq.~\eqref{eq:supp_loglik_lb} to obtain the \textit{\textbf{Hard Vicinal NLL}} (HV-NLL):
	\[
	& \ex_{y\sim p(y)} \left[ \sum_{i=1}^{N} \ind_{\{|y_i-y|\leq\kappa\}} \left( -\log f_{\bm{\theta}}(\bm{x}_{i}^0|y) \right) \right] \label{eq:supp_vicinal_nll}  
	\]
	Replacing $p(y)$ with its \textit{Kernel Density Estimation} (KDE)~\cite{silverman1986density} (bandwidth $\sigma_\delta^2$) yields an approximation of the HV-NLL:
	\[
	& \ex_{y\sim p(y)} \left[ \sum_{i=1}^{N} \ind_{\{|y_i-y|\leq\kappa\}} \left( -\log f_{\bm{\theta}}(\bm{x}_{i}^0|y) \right) \right] \\  
	\approx & \frac{C}{N}\int\left[ \sum_{i=1}^{N} \ind_{\{|y_i-y|\leq\kappa\}} \left(-\log f_{\bm{\theta}}(\bm{x}_{i}^0|y)\right) \right] \\
	& \qquad \cdot \left[ \frac{1}{\sqrt{2\pi}\sigma_\delta} \sum_{j=1}^N \exp\left(-\frac{(y-y_j)^2}{2\sigma_\delta^2}\right) \right] \dee y\\
	& \text{(Let $\delta\triangleq y-y_j$, where $\delta\sim\distNorm(0,\sigma_\delta^2)$)} \\
	= & \frac{C}{N}\int \sum_{i=1}^N\sum_{j=1}^N\left[ W_h \cdot \left( -\log f_{\bm{\theta}}(\bm{x}^0_i|y_j+\delta) \right)  \cdot\frac{e^{-\frac{\delta^2}{2\sigma_\delta^2}}}{\sqrt{2\pi}\sigma_\delta} \right]\dee \delta \\
	= & \frac{C}{N} \sum_{i=1}^N\sum_{j=1}^N \left\{ \ex_{\delta\sim \distNorm(0,\sigma_\delta^2)}\left[ W_h \cdot ( -\log f_{\bm{\theta}}(\bm{x}_i^0|y_j+\delta) ) \right]\right\}
	\label{eq:supp_vicinal_nll_approx} 
	\]
	where $C$ is a constant and
	\[
	W_h\triangleq\ind_{\{|y_j+\delta-y_i|\leq\kappa\}}.
	\label{eq:supp_W_h}
	\]
	The hard vicinal weight $W_h$ can alternatively be substituted with a soft vicinal weight defined by \cite{ding2023ccgan} as follows:
	\[
	W_s\triangleq e^{-\nu(y_j+\delta-y_i)^2}.
	\label{eq:supp_W_s}
	\]
	The positive hyperparameters $\kappa$, $\nu$, and $\sigma_\delta$ in the above formulas can be determined using guidelines outlined in \cite{ding2023ccgan}. 
	
	Eq.~\eqref{eq:supp_vicinal_nll} is not analytically tractable, so instead of directly optimizing Eq.\eqref{eq:supp_vicinal_nll}, we minimize its upper bound. Following the approach outlined in \cite{ho2020denoising, luo2022understanding, murphy2023probabilistic}, the key step is to derive the variational upper bound of $-\log f_{\bm{\theta}}(\bm{x}_i^0|y_j+\delta)$ in Eq.\eqref{eq:supp_vicinal_nll}, which is expressed as follows:
	\[
	& -\log f_{\bm{\theta}}(\bm{x}_i^0|y_j+\delta) \\
	\leq & \kl{q(\bm{x}_i^T|\bm{x}_i^0,y_j+\delta)}{\distNorm(\bm{0},\bm{I})}\label{eq:supp_prior_matching}   \\
	&+\sum_{t=2}^T\ex_{q(\bm{x}_i^t|\bm{x}_i^0,y_j+\delta)}\left[ \kl{q^{t-1}}{p_{\bm{\theta}}^{t-1}} \right] \label{eq:supp_denoising_matching} \\
	&-\ex_{q(\bm{x}_i^1|\bm{x}_i^0,y_j+\delta)}\log p_{\bm{\theta}}(\bm{x}_i^0|\bm{x}_i^1,y_j+\delta)  \label{eq:supp_reconstruction}
	\]
	where $\kl{\cdot}{\cdot}$ is the Kullback–Leibler divergence, and
	\[
	& q^{t-1}\triangleq q(\bm{x}_i^{t-1}|\bm{x}_i^t,\bm{x}_i^0,y_j+\delta) \\
	& p_{\bm{\theta}}^{t-1}\triangleq p_{\bm{\theta}}(\bm{x}_i^{t-1}|\bm{x}_i^t,y_j+\delta)
	\]
	As noted in \cite{luo2022understanding}, the summation term (Eq.~\eqref{eq:supp_denoising_matching}) dominates the reconstruction term (Eq.~\eqref{eq:supp_reconstruction}), while Eq.~\eqref{eq:supp_prior_matching} is a constant independent of $\bm{\theta}$. Thus, we focus on Eq.~\eqref{eq:supp_denoising_matching}. Since both $q^{t-1}$ and $p_{\bm{\theta}}^{t-1}$ are Gaussian, the KL divergence $\kl{q^{t-1}}{p_{\bm{\theta}}^{t-1}}$ can be computed analytically. Moreover, \cite{luo2022understanding} showed that minimizing the summation term across all noise levels in Eq.~\eqref{eq:supp_denoising_matching} can be approximated by minimizing the expectation over all time steps. Given $\bm{x}^t = \sqrt{\bar{\alpha}_t}\bm{x}^0+\sqrt{1-\bar{\alpha}_t}\bm{\eps}$, we have:
	\[
	& \argmin_{\bm{\theta}} \sum_{t=2}^T\ex_{q(\bm{x}_i^t|\bm{x}_i^0,y_j+\delta)} \left[ \kl{q^{t-1}}{p_{\bm{\theta}}^{t-1}} \right] \\
	\approx & \argmin_{\bm{\theta}} \ex_{\bm{\epsilon}\sim\distNorm(\bm{0},\bm{H}_{y_j+\delta}), t\sim U(2,T)}  \\
	&\quad  \left[ \lambda_t \cdot \left( \hat{\bm{x}}_{\bm{\theta}}^0(\sqrt{\bar{\alpha}_t}\bm{x}^0_i+\sqrt{1-\bar{\alpha}_t}\bm{\epsilon},t,y_j+\delta) -\bm{x}^0_i \right)^\intercal  \right. \\
	& \quad \left. \cdot \bm{H}_{y_j+\delta}^{-1} \cdot \left( \hat{\bm{x}}_{\bm{\theta}}^0(\sqrt{\bar{\alpha}_t}\bm{x}^0_i+\sqrt{1-\bar{\alpha}_t}\bm{\epsilon},t,y_j+\delta) -\bm{x}^0_i \right)  \right] \\
	\triangleq & \argmin_{\bm{\theta}} \ex_{\bm{\epsilon}\sim\distNorm(\bm{0},\bm{H}_{y_j+\delta}), t\sim U(2,T)}  \\
	& \quad \left[ \lambda_t \cdot (\hat{\bm{x}}_{\bm{\theta}}^0 -\bm{x}^0_i)^\intercal \bm{H}_{y_j+\delta}^{-1} (\hat{\bm{x}}_{\bm{\theta}}^0 -\bm{x}^0_i) \right] \label{eq:supp_simplified_KL}
	\]
	where,
	\[
	\lambda_t\triangleq \frac{1}{2\sigma_q^2(t)}\frac{\bar{\alpha}_{t-1}(1-\alpha_t)^2}{(1-\bar{\alpha}_t)^2}
	\]
	Empirical studies by Ho et al.~\cite{ho2020denoising} demonstrate that setting $\lambda_t=1$ consistently yields visually superior samples. Thus, we also adopt $\lambda_t=1$ for training our CCDM.
	
	Based on the preceding analysis, we substitute $-\log f_{\bm{\theta}}(\bm{x}_i^0|y_j+\delta)$ in Eq.\eqref{eq:supp_vicinal_nll} with the objective function from Eq.\eqref{eq:supp_simplified_KL}, omit the constant $C$, and let the time step $t$ start from 1. This yields the simplified loss for training U-Net, which we call the \textit{\textbf{Hard Vicinal Image Denoising Loss}} (HVIDL). It is defined as follows:
	\[
	\mathcal{L}(\bm{\theta}) = & -\frac{1}{N} \sum_{i=1}^N\sum_{j=1}^N \left\{ \ex_{\delta\sim \distNorm(0,\sigma_\delta^2), \bm{\epsilon}\sim\distNorm(\bm{0},\bm{H}_{y_j+\delta}), t\sim U(1,T)}  \vphantom{\bm{x}^0_i} \right. \\
	&\left. \vphantom{\ex_{\delta\sim \distNorm(0,\sigma_\delta^2)}} \left[ W_h \cdot (\hat{\bm{x}}_{\bm{\theta}}^0 -\bm{x}^0_i)^\intercal \bm{H}_{y_j+\delta}^{-1} (\hat{\bm{x}}_{\bm{\theta}}^0 -\bm{x}^0_i) \right] \right\}. 
	\label{eq:supp_final_loss}
	\]
	where $W_h$ has already been defined in Eq.~\eqref{eq:supp_W_h}.

	\section{Rule of Thumb for Selecting Hyperparameters}\label{sec:supp_algorithms}
	
	Additionally, as noted in Remark \ref{rmk:vic_param_sel}, we follow the rule of thumb from Remark 3 of \cite{ding2023ccgan} to determine the vicinity-related hyperparameters: $\kappa$, $\nu$, and $\sigma_\delta$. For clarity, we restate this rule below. CcGAN begins by normalizing regression labels into the range $[0,1]$, and hyperparameters are selected based on these normalized labels. The hyperparameter $\sigma_\delta$ is computed using a rule-of-thumb formula for bandwidth selection in \textit{Kernel Density Estimation} (KDE) \cite{silverman1986density}:
	\[
	\sigma_\delta=\left(\frac{4\hat{\sigma}^5_{y}}{3N}\right)^{1/5}, \label{eq:supp_kde_bandwidth_formula}
	\]
	where $\hat{\sigma}_y$ is the sample standard deviation of the normalized training labels. Let $\kappa_{\text{base}}$ be defined as:
	\[
	\kappa_{\text{base}}=\max\left(y_{[2]}-y_{[1]}, y_{[3]}-y_{[2]}, \dots, y_{[\tilde{N}]}-y_{[\tilde{N}-1]} \right)
	\]
	where $y_{[l]}$ denotes the $l$-th smallest normalized distinct label, and $\tilde{N}$ is the number of distinct normalized labels. The parameter $\kappa$ in Eq. \eqref{eq:supp_W_h} for hard vicinity is set as:
	\[
	\kappa = m_{\kappa} \kappa_{\text{base}}, \label{eq:supp_rule_kappa}
	\]
	where $m_{\kappa}\in \mathbb{Z}^+$ indicates half the minimum number of neighboring labels used for estimating $p_r(\bm{x}|y)$ give a $y$ (e.g., $m_{\kappa} = 1$ means using two neighbors, one on each side). In practice, $m_{\kappa}$ is typically 1 or 2, but in extreme cases (e.g., Steering Angle experiments), it may be increased (e.g., $m_{\kappa}=5$). Lastly, the soft vicinal weight parameter $\nu$ in Eq. \eqref{eq:supp_W_s} is determined by:
	\[
	\nu = {1}/{\kappa^2}, \label{eq:supp_rule_nu}
	\]
	which has consistently performed well in \cite{ding2023ccgan}.

	\section{More Details for The Main Experiment}
	
	\subsection{Detailed Training Setups}\label{sec:supp_detailed_train_setups}
	
	Adopting the methodology outlined by Ding et al.~\cite{ding2021ccgan, ding2023ccgan}, regression labels $y$ are normalized to the interval $[0,1]$ when training CCGM models. For class-conditional models, these labels are discretized into distinct classes. The hyperparameters $\kappa$ and $\sigma^2_\delta$ for hard vicinity are chosen based on the guideline in Remark 3 of \cite{ding2023ccgan}. Detailed training configurations for the main results in Table~\ref{tab:main_results} are available in our code repository.
	
	\subsection{Detailed Evaluation Setups}\label{sec:supp_detailed_test_setups}
	
	The evaluation setups follow the methodologies in \cite{ding2023ccgan, ding2024turning}. For each dataset, we define $m_c$ distinct \textbf{evaluation centers} within the range of $y$, designated as $y^{(1)}_e<y^{(2)}_e<...<y^{(m_c)}_e$. Subsequently, each model generates $N_c^g$ synthetic images per center, resulting in $m_c \times N_c^g$ fake image-label pairs. The process for generating these fake samples is detailed as follows: \textbf{(1) RC-49} includes 899 distinct yaw angles, each associated with 49 images representing different chair types. These angles serve as evaluation centers, with each method generating 200 images per angle. Thus, the evaluation includes 179,800 fake image-label pairs. \textbf{(2) UTKFace} includes facial images of individuals aged 1 to 60. These 60 distinct ages are used as evaluation centers, with each method generating 1,000 images per age, resulting in 60,000 fake image-label pairs. \textbf{(3) Cell-200} contains grayscale cell images labeled with 200 distinct cell counts (1 to 200), which are used as evaluation centers. Each method generates 1,000 synthetic images per cell count, resulting in 200,000 fake image-label pairs. \textbf{(4) Steering Angle} comprises 12,271 RGB images labeled with 1,904 distinct steering angles ranging from $-80^\circ$ to $80^\circ$. We define 2,000 evenly spaced evaluation centers within this range and generate 50 images per center for each method. Unlike other datasets, these centers do not align with the dataset’s distinct angles, resulting in 100,000 fake image-label pairs.
	
	We evaluate the synthetic image-label pairs using four metrics: \textbf{(1) SFID:} FID fails to assess conditional generative models due to its lack of conditional information. To address this, \textbf{Intra-FID} \cite{zhang2019self} was introduced, calculating FID between real and fake images at each evaluation center and reporting the average (and \textit{often the standard deviation to reflect model performance across different labels}) across centers. However, Intra-FID struggles in CCGM tasks, such as the Steering Angle experiment, where some centers lack sufficient real images for FID computation. To overcome this, Ding et al.~\cite{ding2023ccgan} proposed \textbf{SFID}, which computes FID within intervals of radius $r_{SFID}$ around each center and averages these values. Lower SFID scores are preferred, and when $r_{SFID}=0$, SFID reduces to Intra-FID. For RC-49, UTKFace, and Cell-200, we set $r_{SFID}=0$, while for Steering Angle, $r_{SFID}=2^\circ$. \textbf{(2) NIQE} evaluates the visual quality of fake images using real images as reference, with lower scores preferred. Similar to SFID, NIQE is computed for each evaluation center, and the final score is the average of these values, with the standard deviation reported across all centers. \textbf{(3) Diversity} measures the intra-label variety of generated images, with higher values indicating greater diversity. To compute it, we use class labels assigned to training images: each image is tagged with one of $K$ classes alongside its regression label. A pre-trained classification CNN predicts the class labels of fake images at each evaluation center, and the Diversity score for each center is determined by the entropy of these predicted labels. The overall Diversity score is the average across all centers, with the standard deviation reported. In our experiments, RC-49 uses 49 chair types, UTKFace uses 5 races, and Steering Angle uses 5 scenes as class labels. Note that Cell-200 lacks class labels, making Diversity inapplicable. \textbf{(4) Label Score} measures label consistency, with smaller values indicating better consistency. A pre-trained regression-oriented ResNet-34 predicts the regression labels of all fake images, which are then compared to the assigned conditional labels. Label Score is defined as the \textit{Mean Absolute Error} (MAE) between predicted and assigned labels, averaged across all fake images, with the standard deviation reported. \textit{Note: For Table \ref{tab:main_results}, Label Score is computed over all fake images, while for Fig.~\ref{fig:line_graphs_rc49}, it is calculated at each of the 899 evaluation angles in RC-49.}

	\begin{table*}[!h] 
		\centering
		\caption{Detailed Implementation Setups for Compared Methods in Table I (Part I).}
		\begin{adjustbox}{width=1\textwidth}
			\begin{tabular}{ccl}
				\toprule
				\textbf{Dataset} & \textbf{Method} & \textbf{Setup} \\
				\midrule
				
				\multirow{8}[17]{*}{\begin{tabular}[c]{@{}c@{}} \textbf{RC-49} \\ {(64$\times$64)}\end{tabular}} & ReACGAN & \begin{tabular}[c]{@{}l@{}} big resnet, hinge loss, steps=$30K$, batch size=256, $\text{lr}_D=4\times 10^{-4}$, $\text{lr}_G=1\times 10^{-4}$\end{tabular}   \\
				\cline{2-3}\specialrule{0em}{1pt}{1pt}
				
				& ADCGAN & \begin{tabular}[c]{@{}l@{}} big resnet, hinge loss, steps=$20K$, batch size=128, update $D$ twice per step, $\text{lr}_D=2\times 10^{-4}$, $\text{lr}_G=5\times 10^{-5}$\end{tabular} \\
				\cline{2-3}\specialrule{0em}{1pt}{1pt}
				
				& ADM-G & \begin{tabular}[c]{@{}l@{}} Classifier: steps=$20K$, batch size=128, lr=$3\times 10^{-4}$ \\ Diffusion: steps=50K, batch size=32, lr=$1\times 10^{-4}$, diffusion steps=$4K$  \end{tabular}   \\
				\cline{2-3}\specialrule{0em}{1pt}{1pt}
				
				& CFG & steps=$50K$, lr=$10^{-4}$, batch size=128, time steps (train/sampling)=1000/100 \\
				\cline{2-3}\specialrule{0em}{1pt}{1pt}
				
				&  CcGAN & \begin{tabular}[c]{@{}l@{}} SAGAN, hinge loss, SVDL+ILI, $\sigma=0.047$, $\nu=50625$, use DiffAugment, steps=$30K$, batch size=256, lr=$10^{-4}$, \\ update $D$ twice per step\end{tabular} \\
				\cline{2-3}\specialrule{0em}{1pt}{1pt}
				
				& Dual-NDA & \begin{tabular}[c]{@{}l@{}} SAGAN, hinge loss, SVDL+ILI, $\sigma=0.047$, $\nu=50625$, $\kappa=0.0044$, use DiffAugment, steps=$40K$, Dual-NDA starts at $30K$,  \\ batch size=256, lr=$10^{-4}$, update $D$ twice per step, 
					$\lambda_1=0.15$, $q_1=0.9$, $\lambda_2=0.15$, $q_2=0.5$, $N_\text{II}=18K$ \end{tabular} \\
				\cline{2-3}\specialrule{0em}{1pt}{1pt}
				
				& CcDPM & \begin{tabular}[c]{@{}l@{}}  $\bm{\epsilon}$-prediction, steps=$50K$, soft vicinity, $\sigma=0.047$, $m_\kappa=2$, $\nu=50625.00$, $p_{\text{drop}}=0.1$, $\gamma=1.5$, timesteps in training $T=1000$, \\ timesteps in sampling $T^\prime=250$, batch size=128, lr=$10^{-4}$ \end{tabular} \\
				\cline{2-3}\specialrule{0em}{1pt}{1pt}
				
				& CCDM & \begin{tabular}[c]{@{}l@{}} $\bm{x}^0$-prediction, steps=$50K$, hard vicinity, $\sigma=0.047$, $m_\kappa=2$, $\kappa=0.0044$, $p_{\text{drop}}=0.1$, $\gamma=1.5$, timesteps in training $T=1000$,  \\ timesteps in sampling $T^\prime=250$, batch size=128, lr=$10^{-4}$ \end{tabular} \\ 
				\midrule

				\multirow{8}[17]{*}{\begin{tabular}[c]{@{}c@{}} \textbf{UTKFace} \\ {(64$\times$64)}\end{tabular}} & ReACGAN & \begin{tabular}[c]{@{}l@{}} big resnet, hinge loss, steps=$40K$, batch size=256, update $D$ twice per step, $\text{lr}_D=2\times 10^{-4}$, $\text{lr}_G=5\times 10^{-5}$\end{tabular}   \\
				\cline{2-3}\specialrule{0em}{1pt}{1pt}
				
				& ADCGAN & \begin{tabular}[c]{@{}l@{}} big resnet, hinge loss, steps=$20K$, batch size=128, update $D$ twice per step, $\text{lr}_D=2\times 10^{-4}$, $\text{lr}_G=5\times 10^{-5}$\end{tabular} \\
				\cline{2-3}\specialrule{0em}{1pt}{1pt}
				
				& ADM-G & \begin{tabular}[c]{@{}l@{}} Classifier: steps=$20K$, batch size=128, lr=$3\times 10^{-4}$ \\ Diffusion: steps=65K, batch size=64, lr=$1\times 10^{-5}$, diffusion steps=$1K$  \end{tabular}  \\
				\cline{2-3}\specialrule{0em}{1pt}{1pt}
				
				& CFG & steps=$100K$, lr=$10^{-4}$, batch size=1024, time steps (train/sampling)=1000/100 \\
				\cline{2-3}\specialrule{0em}{1pt}{1pt}
				
				&  CcGAN & \begin{tabular}[c]{@{}l@{}} SNGAN, the vanilla loss, SVDL+ILI, $\sigma=0.041$, $\nu=3600$, use DiffAugment, steps=$40K$, batch size=256, lr=$10^{-4}$, \\ update $D$ twice per step\end{tabular} \\
				\cline{2-3}\specialrule{0em}{1pt}{1pt}
				
				& Dual-NDA & \begin{tabular}[c]{@{}l@{}} SNGAN, the vanilla loss, SVDL+ILI, $\sigma=0.041$, $\nu=3600$, $\kappa=0.017$, use DiffAugment, steps=$60K$, Dual-NDA starts at $40K$,  \\ batch size=256, lr=$10^{-4}$, update $D$ twice per step, 
					$\lambda_1=0.05$, $q_1=0.9$, $\lambda_2=0.15$, $q_2=0.9$, $N_\text{II}=60K$ ($1K$ per age value)\end{tabular} \\
				\cline{2-3}\specialrule{0em}{1pt}{1pt}
				
				& CcDPM & \begin{tabular}[c]{@{}l@{}}  $\bm{\epsilon}$-prediction, steps=$100K$, soft vicinity, $\sigma=0.041$, $m_\kappa=1$, $\nu=3600$, $p_{\text{drop}}=0.1$, $\gamma=1.5$, timesteps in training $T=1000$,   \\ timesteps in sampling $T^\prime=250$, batch size=128, lr=$10^{-4}$ \end{tabular} \\
				\cline{2-3}\specialrule{0em}{1pt}{1pt}
				
				& CCDM & \begin{tabular}[c]{@{}l@{}} $\bm{x}^0$-prediction, steps=$100K$, hard vicinity, $\sigma=0.041$, $m_\kappa=1$, $\kappa=0.017$, $p_{\text{drop}}=0.1$, $\gamma=1.5$, timesteps in training $T=1000$,  \\ timesteps in sampling $T^\prime=250$, batch size=128, lr=$10^{-4}$ \end{tabular} \\ 
				\midrule

				\multirow{8}[21]{*}{\begin{tabular}[c]{@{}c@{}} \textbf{Steering Angle} \\ {(64$\times$64)}\end{tabular}} & ReACGAN & \begin{tabular}[c]{@{}l@{}} big resnet, hinge loss, steps=$20K$, batch size=256, update $D$ twice per step, $\text{lr}_D=2\times 10^{-4}$, $\text{lr}_G=5\times 10^{-5}$\end{tabular}   \\
				\cline{2-3}\specialrule{0em}{1pt}{1pt}
				
				& ADCGAN & \begin{tabular}[c]{@{}l@{}} big resnet, hinge loss, steps=$20K$, batch size=128, update $D$ twice per step, $\text{lr}_D=2\times 10^{-4}$, $\text{lr}_G=5\times 10^{-5}$\end{tabular} \\
				\cline{2-3}\specialrule{0em}{1pt}{1pt}
				
				& ADM-G & \begin{tabular}[c]{@{}l@{}} Classifier: steps=$20K$, batch size=128, lr=$3\times 10^{-4}$ \\ Diffusion: steps=50K, batch size=32, lr=$3\times 10^{-4}$, diffusion steps=$4K$  \end{tabular}   \\
				\cline{2-3}\specialrule{0em}{1pt}{1pt}
				
				& CFG & steps=$80K$, lr=$10^{-4}$, batch size=128, time steps (train/sampling)=1000/100 \\
				\cline{2-3}\specialrule{0em}{1pt}{1pt}
				
				&  CcGAN & \begin{tabular}[c]{@{}l@{}} SAGAN, the hinge loss, SVDL+ILI, $\sigma=0.029$, $\nu=1000.438$, use DiffAugment, steps=$20K$, batch size=512, lr=$10^{-4}$, \\ update $D$ twice per step\end{tabular} \\
				\cline{2-3}\specialrule{0em}{1pt}{1pt}
				
				& Dual-NDA & \begin{tabular}[c]{@{}l@{}} SAGAN, the hinge loss, SVDL+ILI, $\sigma=0.029$, $\nu=1000.438$, $\kappa=0.032$, use DiffAugment, steps=$20K$, Dual-NDA starts at 0, \\ batch size=512, lr=$10^{-4}$, update $D$ twice per step, 
					$\lambda_1=0.1$, $q_1=0.5$, $\lambda_2=0.2$, $q_2=0.9$, \\ $N_\text{II}=17740$ ($10$ Type II negative images for 1774 distinct training angle values)\end{tabular} \\
				\cline{2-3}\specialrule{0em}{1pt}{1pt}
				
				& CcDPM & \begin{tabular}[c]{@{}l@{}}  $\bm{\epsilon}$-prediction, steps=$50K$, soft vicinity, $\sigma=0.029$, $m_\kappa=5$, $\nu=1000.438$, $p_{\text{drop}}=0.1$, $\gamma=1.5$, timesteps in training $T=1000$, \\ timesteps in sampling $T^\prime=250$, batch size=128, lr=$10^{-4}$ \end{tabular} \\
				\cline{2-3}\specialrule{0em}{1pt}{1pt}
				
				& CCDM & \begin{tabular}[c]{@{}l@{}} $\bm{x}^0$-prediction, steps=$50K$, hard vicinity, $\sigma=0.029$, $m_\kappa=5$, $\kappa=0.032$, $p_{\text{drop}}=0.1$, $\gamma=1.5$, timesteps in training $T=1000$,  \\  timesteps in sampling $T^\prime=250$, batch size=128, lr=$10^{-4}$ \end{tabular} \\ 
				\midrule
				
				\multirow{8}[18]{*}{\begin{tabular}[c]{@{}c@{}} \textbf{Cell-200} \\ {(64$\times$64)}\end{tabular}} & ReACGAN & \begin{tabular}[c]{@{}l@{}} big resnet, hinge loss, steps=$40K$, batch size=256, update $D$ twice per step, $\text{lr}_D=2\times 10^{-4}$, $\text{lr}_G=5\times 10^{-5}$\end{tabular}   \\
				\cline{2-3}\specialrule{0em}{1pt}{1pt}
				
				& ADCGAN & \begin{tabular}[c]{@{}l@{}} big resnet, hinge loss, steps=$20K$, batch size=128, update $D$ twice per step, $\text{lr}_D=2\times 10^{-4}$, $\text{lr}_G=5\times 10^{-5}$\end{tabular} \\
				\cline{2-3}\specialrule{0em}{1pt}{1pt}
				
				& ADM-G & \begin{tabular}[c]{@{}l@{}} Classifier: steps=$20K$, batch size=128, lr=$3\times 10^{-4}$ \\ Diffusion: steps=65K, batch size=64, lr=$1\times 10^{-5}$, diffusion steps=$1K$  \end{tabular}  \\
				\cline{2-3}\specialrule{0em}{1pt}{1pt}
				
				& CFG & steps=$100K$, lr=$10^{-4}$, batch size=1024, time steps (train/sampling)=1000/100 \\
				\cline{2-3}\specialrule{0em}{1pt}{1pt}
				
				&  CcGAN & \begin{tabular}[c]{@{}l@{}} SNGAN, the vanilla loss, SVDL+ILI, $\sigma=0.041$, $\nu=3600$, use DiffAugment, steps=$40K$, batch size=256, lr=$10^{-4}$, \\ update $D$ twice per step\end{tabular} \\
				\cline{2-3}\specialrule{0em}{1pt}{1pt}
				
				& Dual-NDA & \begin{tabular}[c]{@{}l@{}} SNGAN, the vanilla loss, SVDL+ILI, $\sigma=0.041$, $\nu=3600$, $\kappa=0.017$, use DiffAugment, steps=$60K$, Dual-NDA starts at $40K$,  \\ batch size=256, lr=$10^{-4}$, update $D$ twice per step, 
					$\lambda_1=0.05$, $q_1=0.9$, $\lambda_2=0.15$, $q_2=0.9$, $N_\text{II}=60K$ ($1K$ per age value)\end{tabular} \\
				\cline{2-3}\specialrule{0em}{1pt}{1pt}
				
				& CcDPM & \begin{tabular}[c]{@{}l@{}}  $\bm{\epsilon}$-prediction, steps=$50K$, soft vicinity, $\sigma=0.077$, $m_\kappa=2$, $\nu=2500$, $p_{\text{drop}}=0.1$, $\gamma=1.5$, timesteps in training $T=1000$, \\ timesteps in sampling $T^\prime=250$, batch size=128, lr=$5\times 10^{-5}$ \end{tabular} \\
				\cline{2-3}\specialrule{0em}{1pt}{1pt}
				
				& CCDM & \begin{tabular}[c]{@{}l@{}} $\bm{x}^0$-prediction, steps=$50K$, hard vicinity, $\sigma=0.077$, $m_\kappa=2$, $\kappa=0.02$, $p_{\text{drop}}=0.1$, $\gamma=1.5$, timesteps in training $T=1000$,  \\  timesteps in sampling $T^\prime=250$, batch size=128, lr=$5\times 10^{-5}$ \end{tabular} \\ 
				
				\bottomrule
			\end{tabular}%
		\end{adjustbox}
		\label{tab:train_setup1}%
	\end{table*}%

	\begin{table*}[!h] 
		\centering
		\caption{ Detailed Implementation Setups for Compared Methods in Table I (Part II).}
		\begin{adjustbox}{width=1\textwidth}
			\begin{tabular}{ccl}
				\toprule
				\textbf{Dataset} & \textbf{Method} & \textbf{Setup} \\
				\midrule

				\multirow{8}[16]{*}{\begin{tabular}[c]{@{}c@{}} \textbf{UTKFace} \\ {(128$\times$128)}\end{tabular}} & ReACGAN & \begin{tabular}[c]{@{}l@{}} big resnet, hinge loss, steps=$20K$, batch size=128, update $D$ twice per step, $\text{lr}_D=2\times 10^{-4}$, $\text{lr}_G=5\times 10^{-5}$\end{tabular} \\
				\cline{2-3}\specialrule{0em}{1pt}{1pt}
				
				& ADCGAN & \begin{tabular}[c]{@{}l@{}} big resnet, hinge loss, steps=$20K$, batch size=128, update $D$ twice per step, $\text{lr}_D=2\times 10^{-4}$, $\text{lr}_G=5\times 10^{-5}$, use DiffAugment \end{tabular} \\
				\cline{2-3}\specialrule{0em}{1pt}{1pt}
				
				& ADM-G &  \begin{tabular}[c]{@{}l@{}} Classifier: steps=$20K$, batch size=64, lr=$3\times 10^{-4}$ \\ Diffusion: steps=50K, batch size=24, lr=$1\times 10^{-5}$, diffusion steps=$1K$  \end{tabular} \\
				\cline{2-3}\specialrule{0em}{1pt}{1pt}
				
				& CFG & steps=$50K$, lr=$10^{-5}$, batch size=64, time steps (train/sampling)=1000/100 \\
				\cline{2-3}\specialrule{0em}{1pt}{1pt}
				
				& CcGAN & \begin{tabular}[c]{@{}l@{}} SAGAN, the hinge loss, SVDL+ILI, $\sigma=0.041$, $\nu=900$, use DiffAugment, steps=$20K$, batch size=256, lr=$10^{-4}$, \\ update $D$ four times per step\end{tabular} \\
				\cline{2-3}\specialrule{0em}{1pt}{1pt}
				
				& Dual-NDA & \begin{tabular}[c]{@{}l@{}} SAGAN, the hinge loss, SVDL+ILI, $\sigma=0.041$, $\nu=900$, $\kappa=0.033$, use DiffAugment, steps=$22500$, Dual-NDA starts at $20K$, \\ batch size=256, lr=$10^{-4}$, update $D$ four times per step, 
					$\lambda_1=0.05$, $q_1=0.9$, $\lambda_2=0.15$, $q_2=0.9$, $N_\text{II}=60K$ ($1K$ per age value)\end{tabular} \\
				\cline{2-3}\specialrule{0em}{1pt}{1pt}
				
				& CcDPM & \begin{tabular}[c]{@{}l@{}}  $\bm{\epsilon}$-prediction, steps=$200K$, soft vicinity, $\sigma=0.041$, $m_\kappa=1$, $\nu=3600$, $p_{\text{drop}}=0.1$, $\gamma=2.0$, timesteps in training $T=1000$,   \\ timesteps in sampling $T^\prime=150$, batch size=64, lr=$10^{-5}$ \end{tabular} \\
				\cline{2-3}\specialrule{0em}{1pt}{1pt}
				
				& CCDM & \begin{tabular}[c]{@{}l@{}} $\bm{x}^0$-prediction, steps=$200K$, hard vicinity, $\sigma=0.041$, $m_\kappa=1$, $\kappa=0.017$, $p_{\text{drop}}=0.1$, $\gamma=2.0$, timesteps in training $T=1000$,  \\ timesteps in sampling $T^\prime=150$, batch size=64, lr=$10^{-5}$ \end{tabular} \\ 
				\midrule

				\multirow{8}[20]{*}{\begin{tabular}[c]{@{}c@{}} \textbf{Steering Angle} \\ {(128$\times$128)}\end{tabular}} & ReACGAN & \begin{tabular}[c]{@{}l@{}} big resnet, hinge loss, steps=$20K$, batch size=128, update $D$ twice per step, $\text{lr}_D=2\times 10^{-4}$, $\text{lr}_G=5\times 10^{-5}$\end{tabular}   \\
				\cline{2-3}\specialrule{0em}{1pt}{1pt}
				
				& ADCGAN & \begin{tabular}[c]{@{}l@{}} big resnet, hinge loss, steps=$20K$, batch size=128, update $D$ twice per step, $\text{lr}_D=2\times 10^{-4}$, $\text{lr}_G=5\times 10^{-5}$, use DiffAugment \end{tabular} \\
				\cline{2-3}\specialrule{0em}{1pt}{1pt}
				
				& ADM-G & \begin{tabular}[c]{@{}l@{}} Classifier: steps=$20K$, batch size=64, lr=$3\times 10^{-4}$ \\ Diffusion: steps=50K, batch size=24, lr=$1\times 10^{-5}$, diffusion steps=$1K$  \end{tabular}   \\
				\cline{2-3}\specialrule{0em}{1pt}{1pt}
				
				& CFG & steps=$50K$, lr=$10^{-5}$, batch size=64, time steps (train/sampling)=1000/100 \\
				\cline{2-3}\specialrule{0em}{1pt}{1pt}
				
				&  CcGAN & \begin{tabular}[c]{@{}l@{}} SAGAN, the hinge loss, SVDL+ILI, $\sigma=0.029$, $\nu=1000.438$, use DiffAugment, steps=$20K$, batch size=256, lr=$10^{-4}$, \\ update $D$ twice per step\end{tabular} \\
				\cline{2-3}\specialrule{0em}{1pt}{1pt}
				& Dual-NDA & \begin{tabular}[c]{@{}l@{}} SAGAN, the hinge loss, SVDL+ILI, $\sigma=0.029$, $\nu=1000.438$, $\kappa=0.032$, use DiffAugment, steps=$20K$, Dual-NDA starts at $15K$, \\ batch size=256, lr=$10^{-4}$, update $D$ twice per step, 
					$\lambda_1=0.2$, $q_1=0.5$, $\lambda_2=0.3$, $q_2=0.9$, \\ $N_\text{II}=17740$ ($10$ Type II negative images for 1774 distinct training angle values)\end{tabular} \\
				\cline{2-3}\specialrule{0em}{1pt}{1pt}
				
				& CcDPM & \begin{tabular}[c]{@{}l@{}}  $\bm{\epsilon}$-prediction, steps=$200K$, soft vicinity, $\sigma=0.029$, $m_\kappa=5$, $\nu=1000.438$, $p_{\text{drop}}=0.1$, $\gamma=1.5$, timesteps in training $T=1000$, \\ timesteps in sampling $T^\prime=150$, batch size=64, lr=$5\times 10^{-5}$ \end{tabular} \\
				\cline{2-3}\specialrule{0em}{1pt}{1pt}
				
				& CCDM & \begin{tabular}[c]{@{}l@{}} $\bm{x}^0$-prediction, steps=$200K$, hard vicinity, $\sigma=0.041$, $m_\kappa=5$, $\kappa=0.032$, $p_{\text{drop}}=0.1$, $\gamma=1.5$, timesteps in training $T=1000$,  \\  timesteps in sampling $T^\prime=150$, batch size=64, lr=$5\times 10^{-5}$ \end{tabular} \\ 
				\midrule
				
				\multirow{6}[4]{*}{\begin{tabular}[c]{@{}c@{}} \textbf{UTKFace} \\ {(192$\times$192)}\end{tabular}} 
				&  CcGAN & \begin{tabular}[c]{@{}l@{}} SAGAN, the hinge loss, SVDL+ILI, $\sigma=0.029$, $\nu=1000.438$, use DiffAugment, steps=$20K$, batch size=256, lr=$10^{-4}$, \\ update $D$ twice per step\end{tabular} \\
				\cline{2-3}\specialrule{0em}{1pt}{1pt}
				
				& Dual-NDA & \begin{tabular}[c]{@{}l@{}} SAGAN, the hinge loss, SVDL+ILI, $\sigma=0.029$, $\nu=1000.438$, $\kappa=0.032$, use DiffAugment, steps=$20K$, Dual-NDA starts at $15K$, \\ batch size=256, lr=$10^{-4}$, update $D$ twice per step, 
					$\lambda_1=0.2$, $q_1=0.5$, $\lambda_2=0.3$, $q_2=0.9$, \\ $N_\text{II}=17740$ ($10$ Type II negative images for 1774 distinct training angle values)\end{tabular} \\
				\cline{2-3}\specialrule{0em}{1pt}{1pt}
				
				& CcDPM & \begin{tabular}[c]{@{}l@{}}  $\bm{\epsilon}$-prediction, steps=$300K$, soft vicinity, $\sigma=0.041$, $m_\kappa=1$, $\nu=3600$, $p_{\text{drop}}=0.1$, $\gamma=2.0$, timesteps in training $T=1000$,   \\ timesteps in sampling $T^\prime=150$, batch size=64, lr=$10^{-5}$ \end{tabular} \\
				\cline{2-3}\specialrule{0em}{1pt}{1pt}
				
				& CCDM & \begin{tabular}[c]{@{}l@{}} $\bm{x}^0$-prediction, steps=$300K$, hard vicinity, $\sigma=0.041$, $m_\kappa=1$, $\kappa=0.017$, $p_{\text{drop}}=0.1$, $\gamma=2.0$, timesteps in training $T=1000$,  \\ timesteps in sampling $T^\prime=150$, batch size=64, lr=$10^{-5}$ \end{tabular} \\

				\bottomrule
			\end{tabular}%
		\end{adjustbox}
		\label{tab:train_setup2}%
	\end{table*}%

	\begin{table*}[!h] 
		\centering
		\caption{ Detailed Implementation Setups for DMD2-M in Table II.}
			\begin{tabular}{cl}
				\toprule
				\textbf{Dataset} &  \textbf{Setup} \\
				\midrule
				
				RC-49 & \begin{tabular}[c]{@{}l@{}} SNGAN, hinge loss, steps=$50K$, batch size=128, $m_\kappa=0$, $\kappa=0$, update $D$ twice per step,\\ $w_D=10.0$, $w_G=1.0$, $\text{lr}_D=10^{-4}$, $\text{lr}_G=10^{-4}$, DiffAugment \end{tabular} \\			
				\midrule
				
				Steering Angle & \begin{tabular}[c]{@{}l@{}} SNGAN, hinge loss, steps=$200K$, batch size=128, $m_\kappa=1$, $\kappa=0.011$, update $D$ twice per step,\\ $w_D=10.0$, $w_G=1.0$, $\text{lr}_D=10^{-4}$, $\text{lr}_G=10^{-4}$, DiffAugment \end{tabular}   \\
				\bottomrule
			\end{tabular}%
		\label{tab:train_setup3}%
	\end{table*}%

	\section{More Details for Ablation Study} \label{sec:supp_detailed_ablation}
	
	Unless specified, all ablation studies in Section \ref{sec:ablation} use the training and sampling setups from our main study, with adjustments only to the specific parameter being investigated.
	
\begin{figure*}[!htbp]
		\centering
		\includegraphics[width=0.6\linewidth]{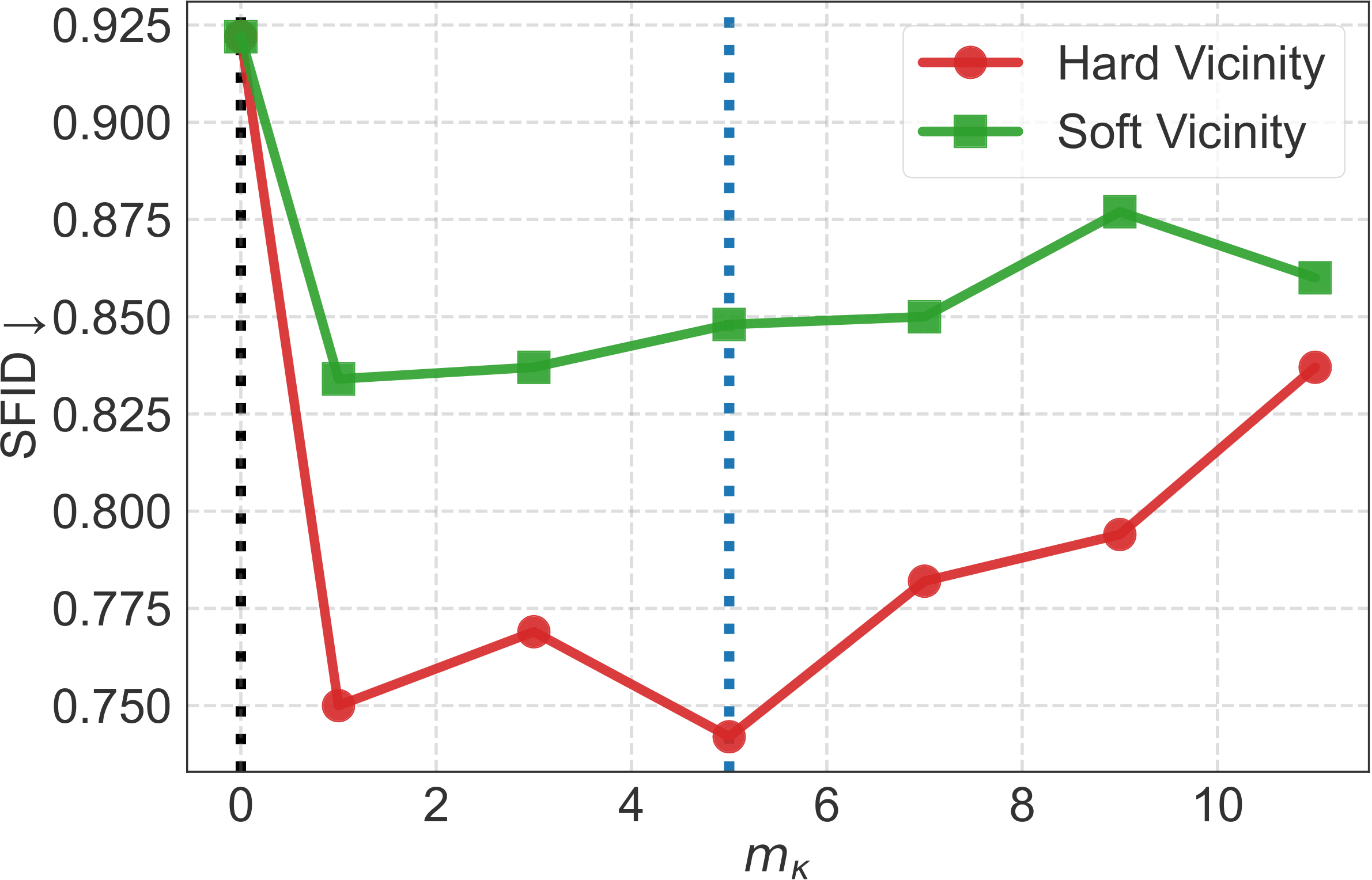} 
		\caption{\textbf{Effect of $m_\kappa$ on CCDM's performance} The blue dashed line indicates the $m_\kappa$ chosen using the rule of thumb from \cite{ding2023ccgan}, while the black dashed line represents no vicinity.} 
		\label{fig:supp_line_graphs_effect_mkappa_sa64}
	\end{figure*}

	To investigate the impact of $m_{\kappa}$ in Eq. \eqref{eq:supp_rule_kappa} and vicinity types on CCDM performance, we conducted an ablation study using Steering Angle ($64\times 64$), with results in Table \ref{tab:supp_ab_effect_kappa_base_sa64} and Fig. \ref{fig:supp_line_graphs_effect_mkappa_sa64}. Following the rule-of-thumb formula in Eq. \eqref{eq:supp_kde_bandwidth_formula}, we computed $\sigma_\delta$ from normalized training labels, resulting in $\sigma_\delta=0.0294$. For each vicinity type, we varied $m_{\kappa}$ from 0 to 11, calculating $\kappa$ and $\nu$ using Eqs. \eqref{eq:supp_rule_kappa} and \eqref{eq:supp_rule_nu}. Key observations include: \textbf{(1)} Our setup with hard vicinity at $m_{\kappa}=5$ yields the best performance. \textbf{(2)} With hard vicinity, SFID accelerates significantly when $m_\kappa>5$ (i.e., $\kappa>0.0316$). \textbf{(3)} With a soft vicinity, the minimum SFID is achieved at $m_\kappa=1$, corresponding to the smallest value of $m_\kappa$. As $m_\kappa$	increases, a clear upward trend in SFID is observed, indicating a gradual degradation in performance. \textbf{(4)} The hard vicinity consistently outperforms the soft vicinity across all $m_{\kappa}$ values. These findings support the use of hard vicinity and confirm that the heuristic for determining $\kappa$ and $\nu$ from CcGANs \cite{ding2023ccgan} remains effective as long as $m_{\kappa}$ is not excessively high.
	
	\begin{table*}[!htbp]
		\centering
		\caption{\textbf{Performance Evaluation of CCDM with Varied Vicinity Setups in Steering Angle ($64\times 64$) Experiments.} The KDE bandwidth is selected based on Eq.~\eqref{eq:supp_kde_bandwidth_formula}, yielding $\sigma_\delta=0.0294$. Our proposed configuration is indicated with an asterisk (*). }
			\begin{tabular}{c|ccc|cccc}
				\toprule
				\begin{tabular}[c]{@{}c@{}} \textbf{Vicinity} \\ \textbf{Type}\end{tabular} & $m_{\kappa}$ & $\kappa$ & $\nu$ &  \textbf{SFID} $\downarrow$  & \textbf{NIQE} $\downarrow$  & \textbf{Diversity} $\uparrow$ & \textbf{Label Score} $\downarrow$  \\
				\midrule
				\begin{tabular}[c]{@{}c@{}} \textbf{No} \\ \textbf{Vicinity}\end{tabular} &  0  &  NA  &  NA  &  0.922  &  1.758  &  1.127  & 7.376 \\
				\cdashline{1-8}
				\specialrule{0em}{1pt}{1pt}
				
				\multirow{6}[0]{*}{\textbf{Hard}*} & 1 &  0.0063  &  NA  & 0.750 & 1.785  & 1.104 & 5.870 \\
				& 3 &  0.0190  &  NA   & 0.769 & 1.785  & 1.125  & 6.252 \\
				& 5* &  0.0316  &  NA   & 0.742 & 1.778  & 1.088  & 5.823 \\
				& 7 &  0.0441  &  NA   & 0.782 & 1.784  & 1.100  & 6.042 \\
				& 9 &  0.0567  &  NA   & 0.794 & 1.783  & 1.102  & 6.016 \\
				& 11 & 0.0693  &  NA   & 0.837 & 1.783  & 1.102  & 6.134 \\
				\cdashline{1-8}
				\specialrule{0em}{1pt}{1pt}
				
				\multirow{6}[0]{*}{\textbf{Soft}} & 1 &  0.0063  &  25195.26 & 0.834 & 1.790  & 1.129  & 6.404 \\
				& 3 &  0.0190  &  2770.08   & 0.837 & 1.781  & 1.121  & 6.361 \\
				& 5 &  0.0316  &  1001.44   & 0.848 & 1.773  & 1.097  & 6.652 \\
				& 7 &  0.0441  &  514.19   & 0.850 & 1.780  & 1.119  & 6.553 \\
				& 9 &  0.0567  &  311.05   & 0.877 & 1.782  & 1.111  & 6.758 \\
				& 11 & 0.0693  &  208.23   & 0.860 & 1.777  & 1.109  & 6.665 \\
				\bottomrule
			\end{tabular}%
		\label{tab:supp_ab_effect_kappa_base_sa64}%
	\end{table*}%
	
\end{document}